\documentclass[10pt,twocolumn,letterpaper]{article}

\usepackage{cvpr}      

\usepackage{graphicx}
\usepackage{amsmath}
\usepackage{amssymb}
\usepackage{booktabs}
\usepackage{comment}
\usepackage{multirow}
\usepackage[colorlinks = true,
            urlcolor  = black,
            citecolor = green,
            anchorcolor = blue]{hyperref}
\usepackage[accsupp]{axessibility}  

\newcommand{\ols}[1]{\mskip.5\thinmuskip\overline{\mskip-.5\thinmuskip {#1} \mskip-.5\thinmuskip}\mskip.5\thinmuskip} 
\newcommand{\olsi}[1]{\,\overline{\!{#1}}} 
\makeatletter
\newcommand\closure[1]{
  \tctestifnum{\count@stringtoks{#1}>1} 
  {\ols{#1}} 
  {\olsi{#1}} 
}
\long\def\count@stringtoks#1{\tc@earg\count@toks{\string#1}}
\long\def\count@toks#1{\the\numexpr-1\count@@toks#1.\tc@endcnt}
\long\def\count@@toks#1#2\tc@endcnt{+1\tc@ifempty{#2}{\relax}{\count@@toks#2\tc@endcnt}}
\def\tc@ifempty#1{\tc@testxifx{\expandafter\relax\detokenize{#1}\relax}}
\long\def\tc@earg#1#2{\expandafter#1\expandafter{#2}}
\long\def\tctestifnum#1{\tctestifcon{\ifnum#1\relax}}
\long\def\tctestifcon#1{#1\expandafter\tc@exfirst\else\expandafter\tc@exsecond\fi}
\long\def\tc@testxifx{\tc@earg\tctestifx}
\long\def\tctestifx#1{\tctestifcon{\ifx#1}}
\long\def\tc@exfirst#1#2{#1}
\long\def\tc@exsecond#1#2{#2}
\makeatother

\usepackage[capitalize]{cleveref}
\crefname{section}{Sec.}{Secs.}
\Crefname{section}{Section}{Sections}
\Crefname{table}{Table}{Tables}
\crefname{table}{Tab.}{Tabs.}

\def\cvprPaperID{4474} 
\def\confName{CVPR}
\def\confYear{2023}

\begin{document}

\title{Event-based Video Frame Interpolation with Cross-Modal Asymmetric Bidirectional Motion Fields}

\author{Taewoo Kim, Yujeong Chae, Hyun-Kurl Jang, Kuk-Jin Yoon \\
Korea Advanced Institute of Science and Technology \\
{\tt\small \{intelpro,yujeong,jhg0001,kjyoon\}@kaist.ac.kr}
}
\maketitle

\begin{abstract}
Video Frame Interpolation (VFI) aims to generate intermediate video frames between consecutive input frames. 
Since the event cameras are bio-inspired sensors that only encode brightness changes with a micro-second temporal resolution, several works utilized the event camera to enhance the performance of VFI.
However, existing methods estimate bidirectional inter-frame motion fields with only events or approximations, which can not consider the complex motion in real-world scenarios.
In this paper, we propose a novel event-based VFI framework with cross-modal asymmetric bidirectional motion field estimation.
In detail, our EIF-BiOFNet utilizes each valuable characteristic of the events and images for direct estimation of inter-frame motion fields without any approximation methods.
Moreover, we develop an interactive attention-based frame synthesis network to efficiently leverage the complementary warping-based and synthesis-based features.
Finally, we build a large-scale event-based VFI dataset, ERF-X170FPS, with a high frame rate, extreme motion, and dynamic textures to overcome the limitations of previous event-based VFI datasets.
Extensive experimental results validate that our method shows significant performance improvement over the state-of-the-art VFI methods on various datasets.
Our project pages are available at: \url{https://github.com/intelpro/CBMNet}
\end{abstract}
\vspace{-8pt}

\section{Introduction}
\label{sec:intro}
Video frame interpolation (VFI) is a long-standing problem in computer vision that aims to increase the temporal resolution of videos.
It is widely applied to various fields ranging from SLAM, object tracking, novel view synthesis, frame rate up-conversion, and video enhancement.
Due to its practical usage, many researchers are engaged in enhancing the performance of video frame interpolation.
Recently, numerous deep learning-based VFI methods have been proposed and recorded remarkable performance improvements in various VFI datasets.
Specifically, numerous motion-based VFI methods~\cite{DAIN,BMBC,ABME,bao2019memc,vfiformer,superslowmo,gui2020featureflow} are proposed thanks to the recent advance in motion estimation algorithms~\cite{teed2020raft,Sun2018PWC-Net,loopnet,luo2021upflow,jonschkowski2020matters,jiang2021learning}.
For the inter-frame motion field estimation, the previous works~\cite{DAIN,BMBC,superslowmo} estimate the optical flows between consecutive frames and approximate intermediate motion fields~\cite{qvi,superslowmo,BMBC} using linear~\cite{BMBC,superslowmo} or quadratic~\cite{qvi} approximation assumptions.
These methods often estimate the inaccurate inter-frame motion fields when the motions between frames are vast or non-linear, adversely affecting the VFI performance.

\begin{figure}[!t]
\centering
\vspace{-3pt}
\includegraphics[width=0.8\linewidth]{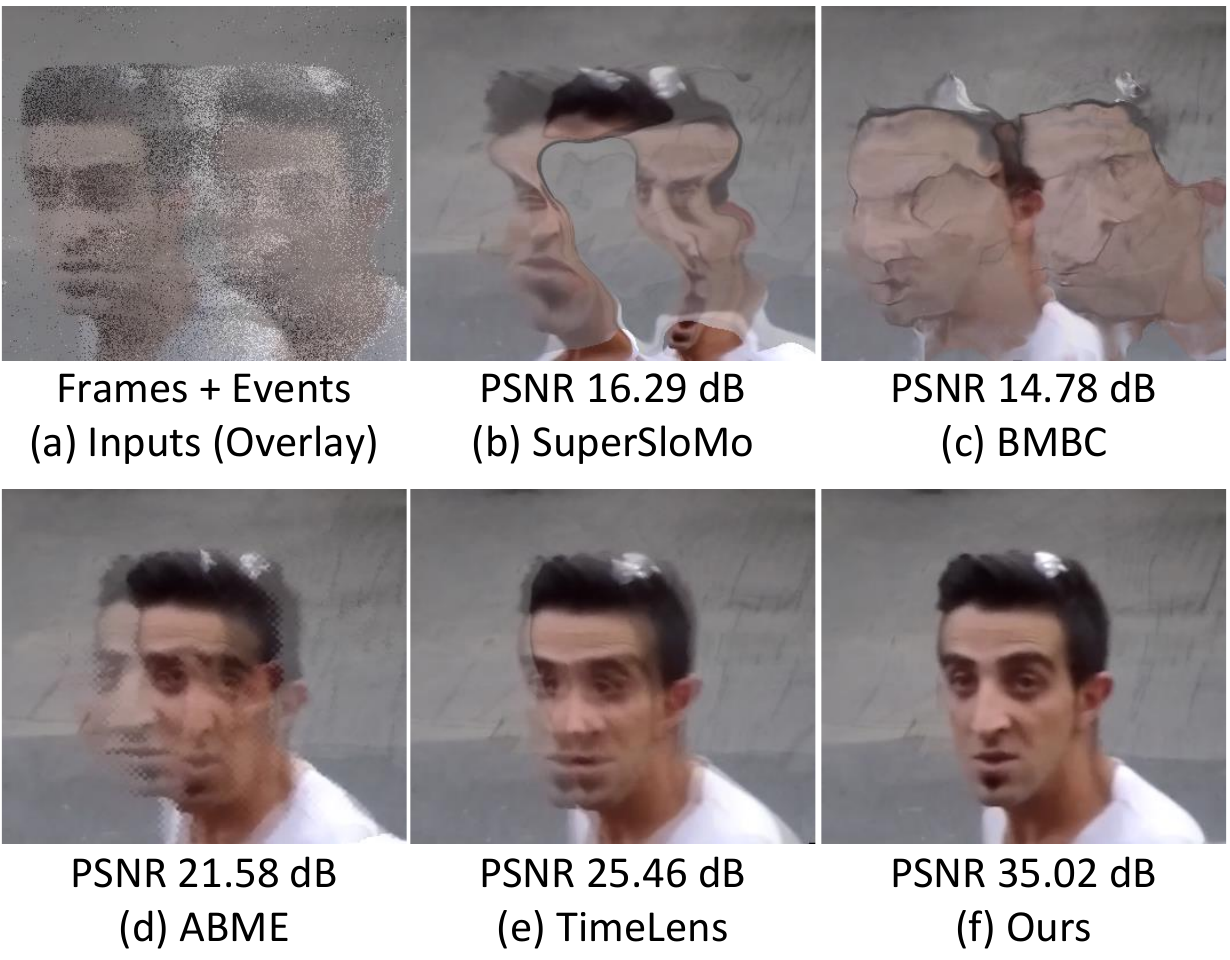} 
\vspace{-3mm}
\captionsetup{font=small}
\caption{Qualitative comparison on the warped frame of inter-frame motion fields. 
(b) and (c) estimate symmetrical inter-frame motion fields. (d) and (e) estimate asymmetric motion fields using only images and events, respectively. (f) Ours shows the best results using cross-modal asymmetric bidirectional motion fields.
}
\label{fig:cover_fig}
\vspace{-4mm}
\end{figure}

Event cameras, novel bio-inspired sensors, can capture blind motions between frames with high temporal resolutions since they asynchronously report the brightness changes of the pixels at the micro-second level. 
Therefore, recent event-based VFI methods~\cite{timelens, timelens++, ledvdi, wevi, timereplayer, a2of} have tried to leverage the advantages of the events to estimate the inter-frame motion field.
However, they utilized only the stream of events to estimate motion fields~\cite{timelens, timereplayer} or used approximation methods~\cite{wevi,timelens++,a2of}, resulting in sub-optimal motion field estimation.
The first reason is that the nature of events is sparse and noisy, and all the brightness changes are not recorded in the stream, leading to inaccurate results.
Second, the approximation-based methods can not completely express the motion model of complex real-world motion fields.
Lastly, the previous works do not fully take advantage of the modality of images providing dense visual information; the previous works do not consider a remarkable distinction between the two modalities, such as simply concatenating two modality features~\cite{timelens++}.

To solve the problem, we propose a novel EIF-BiOFNet(\textbf{E}vent-\textbf{I}mage \textbf{F}usion \textbf{Bi}directional \textbf{O}ptical \textbf{F}low \textbf{Net}work) for directly estimating asymmetrical inter-frame motion fields, elaborately considering the characteristics of the events and frames \textit{without the approximation methods}.
Our EIF-BiOFNet leverage the following properties of events and images in the inter-frame motion field estimation.
Since the events are triggered near the object edges and provide abundant motion trajectory information, the event-level motion fields tend to be accurately estimated near the motion boundaries.
However, it can not give precise results in other areas where the events are not triggered.
On the other hand, the image-based motion fields can utilize dense visual information, but they can not employ the trajectory of motion, unlike the events.
The proposed EIF-BiOFNet elaborately supplements image-level and event-level motion fields to estimate more accurate results.
As a result, the EIF-BiOFNet significantly outperforms previous methods, as shown in Fig.~\ref{fig:cover_fig}.

For the intermediate frame synthesis, recent event-based VFI methods~\cite{timelens,timelens++,a2of,timereplayer} are built on CNNs. However, most CNN-based frame synthesis methods have a weakness in long-range pixel correlation due to the limited receptive field size. To this end, we propose a new Interactive Attention-based frame synthesis network to leverage the complementary \textit{warping}- and \textit{synthesis}-based features. 

Lastly, we propose a novel large-scale ERF-X170FPS dataset with an elaborated beam-splitter setup for event-based VFI research community. 
Existing event-based VFI datasets have several limitations, such as not publicly available train split, low frame rate, and static camera movements. 
Our dataset contains a higher frame rate (170fps), higher resolution (1440$\times$975), and more diverse scenes compared to event-based VFI datasets~\cite{timelens++} where both train and test splits are publicly available.

In summary, our contributions are four fold: (I) We propose a novel EIF-BiOFNet for estimating the asymmetric inter-frame motion fields by elaborately utilizing the cross-modality information.
(II) We propose a new Interactive Attention-based frame synthesis network that efficiently leverages the complementary warping- and synthesis-based features.
(III) We propose a novel large-scale event-based VFI dataset named \textbf{ERF-X170FPS}.
(IV) With these whole setups, we experimentally validate that our method significantly outperforms the SoTA VFI methods on five benchmark datasets, including the ERF-X170FPS dataset. 
Specifically, we recorded \textbf{unprecedented performance improvement} compared to SoTA event-based VFI method~\cite{timelens} by \textbf{7.9dB} (PSNR) in the proposed dataset.

\section{Related Works}
\vspace{-2pt}
\label{sec:Related_works}
\noindent\textbf{Frame-based Video Frame Interpolation.} can be classified into three categories: motion-based approaches~\cite{DAIN,BMBC,ABME,bao2019memc,vfiformer,superslowmo,gui2020featureflow,hu2022many,Danier_2022_CVPR,niklaus2020softmax}, kernel-based approaches~\cite{DAIN,bao2019memc,sepconv,lee2020adacof,cheng2020video,shi2022video} and phase-based approaches~\cite{meyer2015phase,meyer2018phasenet}.
Thanks to the progress of optical flow algorithms~\cite{teed2020raft,Sun2018PWC-Net}, the motion-based approaches are the most actively studied.
This approach generates the intermediate frames by warping pixels using a motion field between frames. 
To be specific, previous works estimated the inter-frame motion fields using the linear~\cite{DAIN,superslowmo,BMBC}  and quadratic approximation.~\cite{qvi,liu2020enhanced} and knowledge distillation~\cite{RIFE,IFRNet}.
BMBC~\cite{BMBC} further proposed a bilateral cost volume layer to enhance the inter-frame motion.
ABME~\cite{ABME} further developed the methods to estimate asymmetric bilateral motion fields using intermediate temporal frame using motion fields from BMBC~\cite{BMBC}.
However, the inter-frame motion field estimated with only frames are still unreliable when the motion of videos is large or severely non-linear.

\noindent\textbf{Event-based Video Frame Interpolation.} Event cameras have the advantages of micro-second level temporal resolution and HDR properties.
Thanks to these novel features, recent event-based research successfully enhanced the quality of the VFI~\cite{kim2021event,Shang_2021_ICCV,sun2022event,ledvdi,xu2021motion,Zhang_2022_CVPR,Song_2022_CVPR}.
In the event-based VFI, TimeLens~\cite{timelens} first proposed a unified video interpolation framework by leveraging both the warping-based and synthesis-based approaches.
Subsequently, some works proposed weakly supervised-based~\cite{wevi} and unsupervised-based~\cite{timereplayer} event-based VFI methods.
In both two works~\cite{timelens,timereplayer}, they estimated inter-frame motion fields through the stream of events only.
However, their motion fields often fail to estimate accurate results due to the sparse nature of events and do not fully utilize the dense visual information of the images.
\cite{timelens++,a2of} mainly focus on the approximation method of the inter-frame motion fields. 
TimeLens++~\cite{timelens++} estimated inter-frame motion with spline approximation and multi-scale fusion method.
Concurrently, A$^2$OF~\cite{a2of} estimated an optical-flow distribution mask with events and utilized it as the approximation weights of inter-frame motion fields.
\textit{Unlike these works, we propose a novel EIF-BiOFNet for directly estimating inter-frame motion fields by fully leveraging cross-modality information without relying on the motion approximation methods.}

\begin{figure*}[!t]
\centering
\vspace{-4pt}
\includegraphics[width=.82\linewidth]{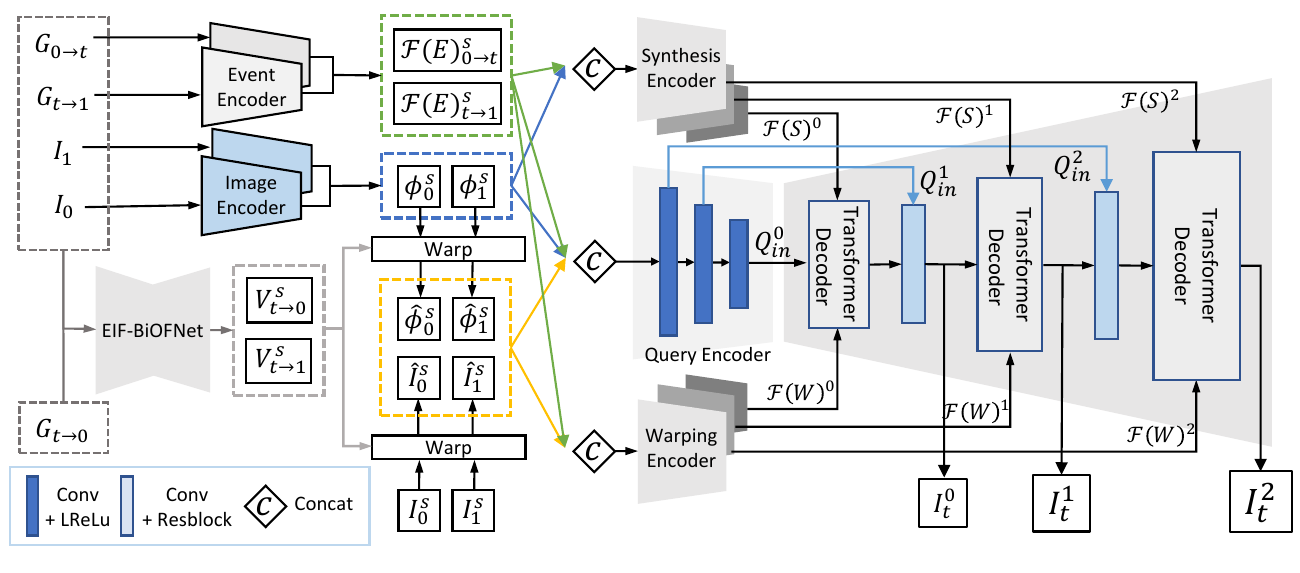} 
\vspace{-14pt}
\caption{The overall architecture of Interactive Attention-based frame synthesis network.}
\label{fig:Arc_frame_synthesis}
\vspace{-8pt}
\end{figure*}

\vspace{-2pt}
\section{Proposed Methods}
\vspace{-2pt}
\subsection{Problem Formulations and Overview}
\vspace{-2pt}
\noindent\textbf{Problem Formulations.} In the event-based VFI works, the first key frame $I_{0}$, the last key frame $I_{1}$ and the inter-frame events $E_{0\rightarrow1}$ are given.
To utilize the inter-frame events $E_{0 \rightarrow 1}$ for video interpolation, we first divide the events $E_{0 \rightarrow 1}$ into left events $E_{0 \rightarrow t}$ and right events $E_{t \rightarrow 1}$ based on an arbitrary time step $t\in\{0, 1\}$ to be interpolated.
To utilize the events as the network inputs, we use the standard event voxel grid~\cite{zhu2019unsupervised}. For all experiments, we set the temporal bins of voxels as 16. We now represent the voxel grid of the events between time steps $t_{a}$ to $t_{b}$ as $G_{t_{a} \rightarrow t_{b}}$.
With two images ($I_{0}$, $I_{1}$) and two events ($E_{0 \rightarrow t}$, $E_{t \rightarrow 1}$) as inputs, we aim to estimate intermediate frame $I_{t}$.
\\
\noindent\textbf{Overview}
The overall structure is illustrated in Fig.~\ref{fig:Arc_frame_synthesis}.
It is composed of two major sub-networks: EIF-BiOFNet and interactive attention-based frame synthesis network.
To set network inputs, we convert two inter-frame events ($E_{0 \rightarrow t}$, $E_{t \rightarrow 1}$) into event voxel grids ($G_{0 \rightarrow t}$, $G_{t \rightarrow 1}$).
In addition, we employ the event reversal method~\cite{timelens} to make $E_{t \rightarrow 0}$ from $E_{0 \rightarrow t}$ and convert it to the voxel grid $G_{t \rightarrow 0}$ for event-only motion field.
Our EIF-BiOFNet receives three event voxel data ($G_{0 \rightarrow t}$, $G_{t \rightarrow 0}$, $G_{t \rightarrow 1}$) and two image data ($I_{0}$, $I_{1}$) as network inputs and directly estimates asymmetric bidirectional motion fields ($V_{t \rightarrow 1}$, $V_{t \rightarrow 0}$).
Using these motion fields, we aim to estimate an intermediate frame $I_{t}$ through an interactive attention-based frame synthesis network.

\vspace{-2pt}
\subsection{EIF-BiOFNet}
\begin{figure*}[!t]
\centering
\vspace{-4pt}
\includegraphics[width=.84\linewidth]{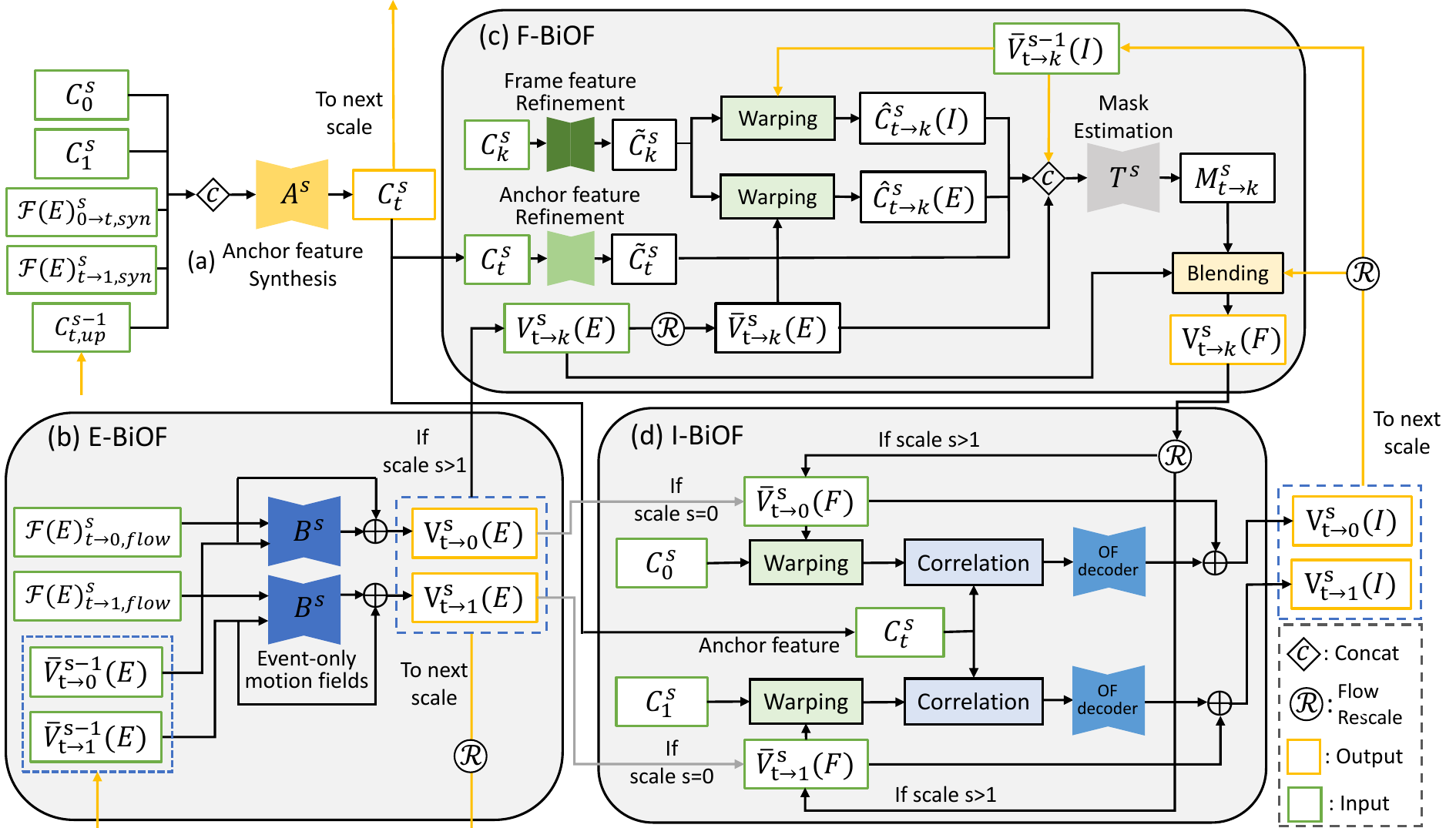} \vspace{-6pt}
\caption{The network architecture of proposed EIF-BiOFNet in scale $s$. For brevity, we only depict unidirectional motion field among the bidirectional motion fields in the F-BiOF module.}
\label{fig:EIF_BiOF}
\vspace{-8pt}
\end{figure*}

\vspace{-2pt}
\noindent\textbf{EIF-BiOFNet Overview.}
The EIF-BiOFNet module in each scale $s$ is illustrated in Fig.\ref{fig:EIF_BiOF}.
The EIF-BiOFNet is composed of a scale-pyramid architecture with scale index $s\in\{0,1,2\}$ to leverage multiple visual scale information.
We extract features through three encoders. First, we extract event features $\{\mathcal{F}(E)^{s}_{t \rightarrow k, flow}\}$ with frame index $k \in \{0, 1\}$ for the event-only motion fields using $G_{t\rightarrow0}$ and $G_{t\rightarrow1}$.
We also extract the left/right event features $\{\mathcal{F}(E)^{s}_{0 \rightarrow t, syn}\}$, $\{\mathcal{F}(E)^{s}_{t \rightarrow 1, syn}\}$ for the anchor frame feature processing using $G_{0 \rightarrow t}$ and $G_{t \rightarrow 1}$ and extract frame feature pyramid $\{C_{k}^{s}\}$.
Then, these cross-modality features pass through four light-weight cascade modules in order: (a) Anchor frame feature pre-processing, (b) E-BiOF, (c) F-BiOF and (d) I-BiOF.
In the first scale ($s$=0), there is no output of I-BiOF at the previous scale, so the outcome of E-BiOF is directly connected to the I-BiOF.
Using the outputs of the I-BiOF at the last scale ($s$=2), EIF-BiOFNet finally produces bilateral motion field pyramids $\{V^{s}_{t \rightarrow 0}\}$, $\{V^{s}_{t \rightarrow 1}\}$.
\\
\noindent \textbf{Anchor Frame Feature Pre-processing.} To perform backward warping from input frames ($I_{0}$, $I_{1}$) to target frame $I_{t}$, we need to estimate the asymmetric motion fields $V_{t \rightarrow 0}$ and $V_{t \rightarrow 1}$.
Without using the approximation methods, we require the intermediate frame $I_{t}$ to estimate $V_{t \rightarrow 0}$ and $V_{t \rightarrow 1}$.
However, $I_{t}$ is unavailable as network inputs.
To solve the aforementioned problem, the recent method~\cite{ABME} generated a temporal ``anchor'' frame $I_{t}$ using existing motion field estimation methods~\cite{BMBC}, then performed correlation between input frames and the temporal anchor frame.
Although they do not rely on the approximation method, the correlation with an incorrectly estimated anchor frame is likely to propagate the error to the motion fields.

In the case of using the events, the ``synthesis-based interpolation'' makes it possible to generate intermediate information between input frames since the events can capture blind motions within $I_{0}$ and $I_{1}$.
Inspired by this fact, we devise a manner to directly synthesize the latent frame features with the same effect as the features obtained from anchor frame $I_{t}$ by fully leveraging the advantages of events.
To this end, we design the light-weight network blocks for anchor frame feature synthesis $A^{s}$ that scale-recurrently synthesize and refine the anchor frame feature $C_{t}^{s}$ using frame features ($C_{0}^{s}$, $C_{1}^{s}$), the left/right event features ($\mathcal{F}(E)^{s}_{0 \rightarrow t, syn}$, $\mathcal{F}(E)^{s}_{t \rightarrow 1, syn}$) and the upsampled anchor frame feature at the previous scale $C_{t,up}^{s-1}$ as in Fig.~\ref{fig:EIF_BiOF}(a).

In this way, the network blocks $A^{s}$ learn to extract the essential information to enable the correlation with the reference frame feature $C_{k}^{s}$ and imitate the anchor frame information.
Since the network blocks learn the feature-level information rather than the image-level information, it allows end-to-end training and lowers the influence of error propagation unlike ~\cite{ABME}.
Through the crucial step, we use these anchor frame features $\{C_{t}^{s}\}$ as the input of F-BiOF and I-BiOF.
\\
\noindent\textbf{E-BiOF Module.}
Although image-based optical flow (OF) enables the estimation of spatially dense motion field, it often fails to estimate accurate results due to the upsampling procedure~\cite{Sun2018PWC-Net,luo2021upflow} and the absence of inter-frame motion information.
Unlike image modality, the events are triggered near object boundaries and provide temporally dense motion trajectory information.
These valuable characteristics of the events enable to complement the image-level motion fields.
To utilize the valuable characteristics of the events, we estimate the event-level motion fields to complement image-level motion fields.
However, for estimating event-level motion fields, it is desirable to minimize spatial down-sampling of event features to preserve spatial information where the events occurred since they are spatially sparse.
In contrast, it is efficient for image-based OF to perform the correlation on a lower spatial scale and then upsampling to reduce the computation.
Therefore, we estimate event-level OF at a larger spatial scale than the image-based OF and perform re-scaling.
In E-BiOF, we estimate residual OF to get $\{V(E)^{s}_{t \rightarrow k}\}$ via decoder unit $B^{s}$ using the event features $\{\mathcal{F}(E)^{s}_{t \rightarrow k,flow}\}$  as in the Fig.~\ref{fig:EIF_BiOF}(b). \\
\noindent\textbf{F-BiOF Module.}
Since EIF-BiOFNet uses the scale-pyramid structure, we perform upsampling on the outputs of I-BiOF at the previous scale and use them as the input of the F-BiOF.
As in \cite{luo2021upflow}, such interpolation-based upsampling of OF can cause adverse effects in motion fields.
Furthermore, it is necessary to utilize inter-frame information from the events efficiently.
For this reason, we additionally use the event-level OF as the input to complement upsampled OF. 
Using these two inputs, we aim to combine the strengths and complement the weak points of each OF.

To accomplish this goal, we selectively merge upsampled OF and the event-level OF by calculating the accurately estimated regions of each optical flow.
As illustrated in Fig.~\ref{fig:EIF_BiOF}(c), we first upsample motion fields $V^{s-1}_{t \rightarrow k}(I)$ from the I-BiOF at the previous scale and get $\ols{V}^{s-1}_{t \rightarrow k}(I)$. 
Then, we refine the frame feature $C_{k}^{s}$ to $\tilde{C}_{k}^{s}$ and refine anchor frame feature $C_{t}^{s}$ to $\tilde{C}_{t}^{s}$ at scale $s$.
After that, the refined frame feature $\tilde{C}_{k}^{s}$ is warped by both rescaled event-based OF $\ols{V}^{s}_{t \rightarrow k}(E)$ and upsampled I-BiOF output $\ols{V}^{s-1}_{t \rightarrow k}(I)$.
In formulation, $\hat{C}^{s}_{t \rightarrow k}(E) = W(\tilde{C}_{k}^{s}, \ols{V}^{s}_{t \rightarrow k}(E))$ and $\hat{C}^{s}_{t \rightarrow k}(I) = W(\tilde{C}_{k}^{s}, \ols{V}^{s-1}_{t \rightarrow k}(I))$ where $W$ denotes backward warping operation.
To take advantage of both motion fields, $\ols{V}(I)^{s-1}_{t \rightarrow k}$ and $V(E)^{s}_{t \rightarrow k}$, we estimate pixel-wise confidence mask $M^{s}_{t \rightarrow k}$ with a refined anchor frame feature $\tilde{C}^{s}_{t}$, two warped frame features ($\hat{C}^{s}_{t \rightarrow k}(E)$, $\hat{C}^{s}_{t \rightarrow k}(I)$) and two motion fields ($\bar{V}^{s-1}_{t \rightarrow k}(I)$, $\bar{V}^{s}_{t \rightarrow k}(E)$).

Intuitively, the confidence masks are calculated by comparing two warped frame features with the refined anchor frame feature.
The network blocks $T^{s}$ for mask estimation increases the confidence value at each pixel of the flows to make the warped feature similar to the anchor frame feature.
  
Finally, we perform blending of two motion fields to take advantage of both two motion fields $\ols{V}^{s-1}_{t \rightarrow k}(I)$ and $V^{s}_{t \rightarrow k}(E)$. The outputs of the F-BiOF are generated as 
\vspace{-2pt}
\begin{gather}
V_{t \rightarrow k}(F)=
M_{t \rightarrow k}^{s}\ols{V}^{s-1}_{t \rightarrow k}(I)
+(1-M^{s}_{t \rightarrow k}) V^{s}_{t \rightarrow k}(E)
\end{gather}
where $V^{s}_{t \rightarrow k}(F)$ are the outputs of F-BiOF.
Since bidirectional motion fields are estimated, the F-BiOF module is repeated twice according to the direction at each scale $s$.
The outputs of F-BiOF are fed into inputs of the I-BiOF, as depicted in Fig.~\ref{fig:EIF_BiOF}(c).

\noindent\textbf{I-BiOF Module.} 
We estimated the bidirectional motion fields via two modules, E-BiOF and F-BiOF. 
However, we did not fully utilize the strength of the images containing dense visual information.
To this end, we propose the I-BiOF module for estimating residual flows of F-BiOF output by leveraging the dense visual information of images.
As illustrated in Fig.~\ref{fig:EIF_BiOF}(d), our I-BiOF module consists of warping, correlation and refinement steps as done in ~\cite{Sun2018PWC-Net}.
We first compute $\{\hat{C}_{t \rightarrow k}^{s}\}$ by backward warping the given frame feature $\{C_{k}^{s}\}$ with the rescaled outputs of F-BiOF, $\{\ols{V}_{t \rightarrow k}^{s}(F)\}$.
Then, the warped frame features $\{\hat{C}_{t \rightarrow k}^{s}\}$ and anchor frame feature $C_{t}^{s}$(previously pre-processed for the correlation in Fig.\ref{fig:EIF_BiOF}(a))  pass through the correlation layer with cost-volume norm~\cite{jonschkowski2020matters}.
Finally, we estimate the residual flows via the decoder unit to yield the outputs of I-BiOF $\{V_{t \rightarrow k}^{s}(I)\}$.
Across cascade modules and multi-scale structure, we gradually reduce the search space of motion fields.

\subsection{Interactive Attention-based Frame Synthesis}
The recent work, TimeLens~\cite{timelens}, proposed the event-based VFI framework by leveraging the benefits of \textit{warping-based} and \textit{synthesis-based} interpolation.
The former works well when the pixel displacement is vast but is weak where the occlusion or brightness inconstancy exists.
In contrast, the latter is robust to the occlusion and not affected by the brightness constancy, but it often fails in the large motion of videos.
TimeLens tried to leverage these complementary advantages through image-level fusion but failed to utilize the rich benefits of feature-level processing.

Recently, the transformer demonstrated its ability to capture long-range dependencies between features in various vision tasks~\cite{vaswani2017attention, liu2021swin,zamir2022restormer,vfiformer,shi2022video}.
To this end, we propose an interactive attention module that effectively combines complementary advantages of warping and synthesis-based interpolation using the \textit{vision transformer} architecture.

As illustrated in Fig.~\ref{fig:Arc_frame_synthesis}, we directly estimate asymmetric bidirectional motion fields, $\{V^{s}_{t\rightarrow0}\}$ and $\{V^{s}_{t\rightarrow1}\}$, with scale index $s \in \{0,1,2\}$ using our EIF-BiOFNet.
Using these inter-frame motion fields, we aim to estimate multi-scale intermediate frames $\{I_t^{s}\}$ using input frames ($I_{0}$, $I_{1}$) and left/right event voxels ($G_{0\rightarrow t}$, $G_{t \rightarrow 1}$).
We first extract the frame feature pyramids $\{\phi_{k}^{s}\}$ and event feature pyramids $\{\mathcal{F}(E)_{0 \rightarrow t}^{s}\}$, $\{\mathcal{F}(E)^{s}_{t \rightarrow 1}\}$ using weight-sharing frames and event encoders.
Then we backward warp the input frames to the warped frame pyramid $\{\hat{I}_{k}^{s}\}$ and the frame features to the warped frame feature pyramid $\{\hat{\phi}_{k}^{s}\}$ where $k\in\{0,1\}$.

\begin{figure}[!t]
  \centering
  \vspace{-4pt}
  \includegraphics[width=0.90\columnwidth]{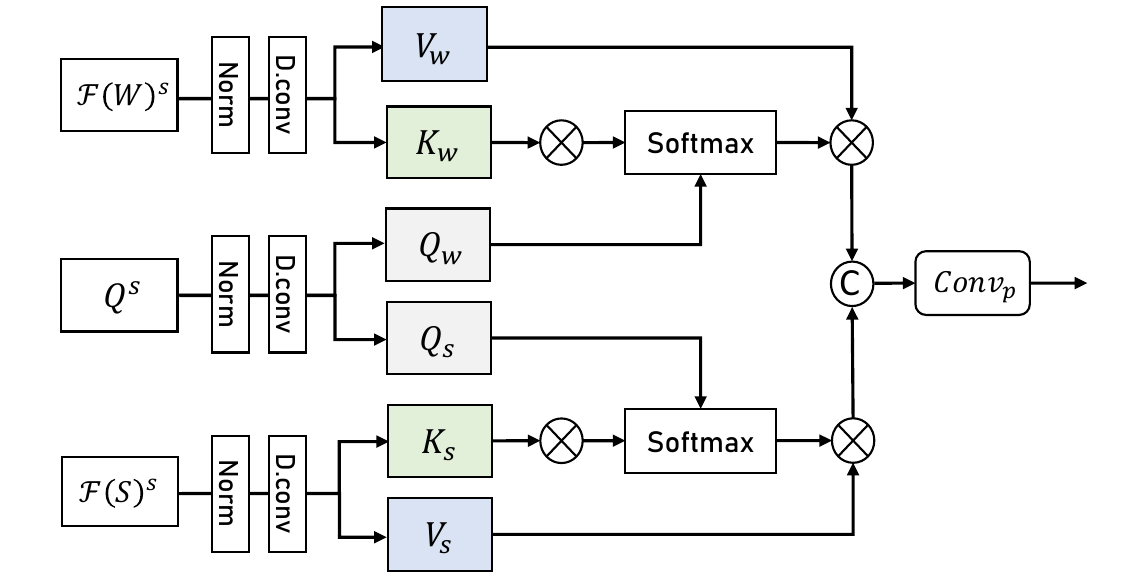}
  \vspace{-2mm}
  \caption{The proposed Interactive Attention Module. © denotes concatenation operation along the channel dimension.}\label{fig:attention} 
\vspace{-4mm}
\end{figure}

\noindent \textbf{Interactive Attention Module.}
Based on the aforementioned inspiration, we devise a hybrid interaction method between \textit{warping-based} and \textit{synthesis-based} features.
To achieve this, we use both the multi-head self-attention and cross-attention operations.
The self-attention mechanism is able to capture long-range dependency between its own feature.
On the other hand, the cross-attention allows the interaction between the input features (\textit{queries}) and the \textit{key}/\textit{value} features.
To this end, we leverage the warping or synthesis features as key/value to enable mutual interaction with queries. 

As in Fig.~\ref{fig:Arc_frame_synthesis}, we build query inputs $Q_{in}^{s}$ by concatenating the whole input features of the last scale and encode the information.
The query input $Q_{in}^{s}$ is connected to the decoder output to make query $Q^{s}$ of the transformer layer input.
For key and value projection, we first encode warping features $\{\mathcal{F}(W)^{s}\}$ utilizing warped frames $\hat{I}_{k}^{s}$, frame features $\hat{\phi}_{k}^{s}$ and left/right event features.
Similarly, we encode synthesis features $\{\mathcal{F}(S)^{s}\}$ using event and frame features.
As shown in Fig.~\ref{fig:attention}, we additionally transform each component (\ie, \textit{query}, \textit{key} and \textit{value}) by applying series of the layer normalization~\cite{ba2016layer} and a depth-wise separable convolution layer.
That is, single query $Q^{s}$ is projected to two queries $Q_{s}$, $Q_{w}$, respectively. 
Also, we project the warped and synthesis feature $\mathcal{F}(W)^{s}$, $\mathcal{F}(S)^{s}$ to two keys $K_{s}$, $K_{w}$ and two values $V_{w}$, $V_{s}$, respectively.
We calculate attentions as follows:
\vspace{-12pt}
\begin{gather}
\mathrm{Attention_{s} = SoftMax} \left( \frac{Q_{s}K_{s}^{T}}{\alpha_{s}} \right) V_{s} \nonumber \\
\mathrm{Attention_{w} = SoftMax} \left( \frac{Q_{w}K_{w}^{T}}{\alpha_{w}} \right) V_{w}
\vspace{-25pt}
\end{gather}
where $\alpha_{s}$ and $\alpha_{w}$ denote learnable scaling parameters to balance the weight of each attention.
For memory efficiency of processing the video frames with high resolution, we compute a cross-covariance matrix ($QK^{T}$) of multi-headed attention across the channel dimension similar to \cite{zamir2022restormer}.
The final attention result $X^{s}$ is generated as:
\vspace{-2pt}
\begin{equation}
X^{s} = Q^{s} + \mathrm{Conv_{p}}[\mathrm{Attention_{s}}, \mathrm{Attention_{k}}]
\vspace{-4pt}
\end{equation}
where $\mathrm{Conv_{p}}$ denotes point-wise conv layer and [] is channel-wise concatenation. 
After the interactive attention blocks, we further refine the output using the self-attention layers.
In this way, we efficiently merge \textit{warping-based} and \textit{synthesis-based} features from the multiple visual scales.
\subsection{Loss Functions}
\vspace{-3pt}
\noindent\textbf{Stage One.} To train the proposed EIF-BiOFNet, we define the photometric loss between GT frame $I_t^{GT}$ and warped frames and edge-aware smoothness loss~\cite{wang2018occlusion} ($L_{smooth}$) as:
\vspace{-2pt}
\begin{small}
\begin{gather}
\vspace{-4pt}
\mathcal{L}_{flow} = \lambda_{1}(\rho(I_{t}^{GT} - W(V_{t \rightarrow 0}, I_{0})) + \rho(I_{t}^{GT} - W(V_{t \rightarrow 1}, I_{1}))) \nonumber \\
+ \lambda_{2}(L_{smooth}(I_{t}^{GT}, V_{t\rightarrow 0}) + L_{smooth}(I_{t}^{GT}, V_{t \rightarrow 1}))
\vspace{-4pt}
\end{gather}
\end{small}
\vspace{-2pt}
where $\rho$ is the charbonnier loss~\cite{charbonnier1994two} and $W$ denotes backward warping operation, and $\lambda_{1},\lambda_{2}$ are balancing parameters between losses, empirically set to $1,10$, respectively.\\
\noindent\textbf{Stage Two.}\ After training stage one, we freeze the weights of the EIF-BiOFNet and train the frame synthesis network using multi-scale loss function as follows:
$\mathcal{L}_{total} = \sum_{s=0}^{2}\lambda_{s}(\rho({I_{t,GT}^{s} - I_{t}^{s}}))$
where $\lambda_{s}$ are set to $\{1, 0.1, 0.1\}$ for each scale.

\begin{table}[!t]
\centering
\caption{A comparison of the datasets for event-based video frame interpolation where train and test sets are both publicly available.}
\vspace{-7pt}
\label{tab:datasets}
\resizebox{\columnwidth}{!}{
\setlength{\tabcolsep}{1pt}
\begin{tabular}{c|ccccc}
\hline
 & RGB cam. & Event cam. & $\mathrm{N}$. seq. & Seq. length & Sync\&align \\ \hline
BS-ERGB~\cite{timelens++} & 970x625, 28fps & 1280x720 & 92 &  100-600 frames & \checkmark \\
ERF-X170FPS & \textbf{1440x975}, \textbf{170fps} & \textbf{1280x720} &  \textbf{140} & \textbf{990 frames} & \checkmark \\ \hline
\end{tabular}
}
\vspace{-10pt}
\end{table}

\begin{table*}[!t]
\centering
\caption{Quantitative evaluation on synthetic event datasets (GoPro and Adobe240fps datasets). The \textbf{bold} and \underline{underlined} denote the best and the second-best performance, respectively. $\dagger$ denotes event-based VFI methods. The same notation and typography are applied to the following tables. Please note that we evaluated all methods based on the middle frame between the input frames.}
\label{tab:synthetic}
\vspace{-5pt}
\scalebox{0.75}{
\setlength{\tabcolsep}{4pt}
\begin{tabular}{c|cccccccccccccccc}
\hline
\multirow{3}{*}{Method} & \multicolumn{8}{c}{GoPro~\cite{nah2017deep}} & \multicolumn{8}{c}{Adobe240fps~\cite{su2017deep}} \\ \cline{2-17} 
 & \multicolumn{2}{c}{7skips} & \multicolumn{2}{c}{15skips} & \multicolumn{2}{c}{23skips} & \multicolumn{2}{c|}{31skips} & \multicolumn{2}{c}{7skips} & \multicolumn{2}{c}{15skips} & \multicolumn{2}{c}{23skips} & \multicolumn{2}{c}{31skips} \\ \cline{2-17} 
 & PSNR & SSIM & PSNR & SSIM & PSNR & SSIM & PSNR & \multicolumn{1}{c|}{SSIM} & PSNR & SSIM & PSNR & SSIM & PSNR & SSIM & PSNR & SSIM \\ \hline
SuperSloMo~\cite{superslowmo} & 27.31 & 0.838 & 22.40 & 0.684 & 20.10 & 0.606 & 18.87 & \multicolumn{1}{c|}{0.566} & 28.10 & 0.879 & 22.69 & 0.727 & 20.26 & 0.639 & 18.72 & 0.588 \\
SepConv~\cite{sepconv} & 27.24 & 0.831 & 21.97 & 0.664  & 19.60 & 0.582 & 18.44 & \multicolumn{1}{c|}{0.543} & 28.13  & 0.875 & 22.40 & 0.713 & 19.97 & 0.622 & 18.51 & 0.572  \\
DAIN~\cite{DAIN} & 26.18 & 0.822 & 22.20 & 0.822 & 20.25 & 0.609 & 19.11 & \multicolumn{1}{c|}{0.571} & 26.63 & 0.859 & 22.29 & 0.719 & 20.08 & 0.634 & 18.67 & 0.585 \\
BMBC~\cite{BMBC} & 26.54 & 0.832 & 21.57 & 0.672 & 19.25 & 0.595 & 18.07 & \multicolumn{1}{c|}{0.559} & 27.18 & 0.874 & 22.19 & 0.722 & 19.82 & 0.637 & 18.36 & 0.590 \\
ABME~\cite{ABME} & 27.79 & 0.845 & 22.71 & 0.690 & 20.51 & 0.616 & 19.31 & \multicolumn{1}{c|}{0.578} & 28.85 & 0.890 & 22.91 & 0.734 & 20.36 & 0.645 & 18.83 & 0.595  \\
RIFE~\cite{RIFE} & 27.60 & 0.837 & 22.57 & 0.681 & 20.37 & 0.606 & 19.16 & \multicolumn{1}{c|}{0.567} & 28.64 & 0.884  & 22.80 & 0.728 & 20.30 & 0.638 & 18.77 & 0.587 \\
IFRNet~\cite{IFRNet} & 27.72 & 0.842 & 22.50 & 0.685 & 20.09 & 0.609 & 18.78 & \multicolumn{1}{c|}{0.573} & 28.66
& 0.886 & 22.72 & 0.729 & 20.21  & 0.642 & 18.67 & 0.593 \\
VFIformer~\cite{vfiformer} & 27.78 & 0.844 & 22.54 & 0.689 & 20.12 & 0.614 & 18.82 & \multicolumn{1}{c|}{0.580} & 28.67 & 0.888 & 22.73 & 0.733 & 20.20 & 0.648 & 18.64 & 0.600 \\ \hline
TimeLens$^{\dagger}$~\cite{timelens} & 34.45 & 0.951 & - & - & - & - & - & \multicolumn{1}{c|}{-} & 34.83 & 0.949 & - & - & - & - & - & - \\
TimeReplayer$^{\dagger}$~\cite{timereplayer} & 33.39 & 0.952 & - & - & - & - & - & \multicolumn{1}{c|}{-} & 33.22 & 0.942 & - & - & - & - & - & - \\
A$^2$OF$^{\dagger}$~\cite{a2of} & 35.95 & 0.967 & - & - & - & - & - & \multicolumn{1}{c|}{-} & 36.21 & 0.957 & - & - & - & - & - & - \\
\hline
Ours & \underline{37.84} & \underline{0.974} & \underline{36.74} & \underline{0.967} & \underline{36.19} & \underline{0.963} & \underline{35.62}  & \multicolumn{1}{c|}{\underline{0.961}} & \underline{37.38} & \underline{0.971}  & \underline{36.38} & \underline{0.966} & \underline{35.45} & \underline{0.961} & \underline{34.88} & \underline{0.958} \\
Ours-Large & \textbf{38.15} & \textbf{0.975} & \textbf{37.05} & \textbf{0.969} & \textbf{36.50} & \textbf{0.965} & \textbf{35.94} & \multicolumn{1}{c|}{\textbf{0.963}} & \textbf{37.76} & \textbf{0.972} & \textbf{36.76} & \textbf{0.968} & \textbf{35.84} & \textbf{0.963} & \textbf{35.29} & \textbf{0.961} \\ \hline
\end{tabular}}
\vspace{-2pt}
\end{table*}

\vspace{-1pt}
\section{ERF-X170FPS Dataset}
\vspace{-1mm}
Behind the remarkable advance of frame-based VFI methods, there are numerous benchmark datasets, such as GOPRO~\cite{nah2017deep}, Adobe240fps~\cite{su2017deep}, Vimeo90K~\cite{xue2019video},
DAVIS~\cite{perazzi2016benchmark}
and XVFI~\cite{sim2021xvfi}. 
In contrast, the public benchmark datasets for event-based VFI research are relatively rare.
The BS-ERGB~\cite{timelens++} is the only publicly available dataset for the training and evaluation of event-based VFI algorithms.
However, the BS-ERGB dataset has several limitations due to the low frame rate (28fps) leading to the large occlusions in the scenes, too many deformable objects (water, fire, popping eggs, \etc.), and static camera movement of the test scenes.
For these reasons, inter-frame motion fields are invalid in many regions in this dataset, which makes it challenging to train and evaluate the performance of the motion-based VFI algorithms.

To tackle the aforementioned problems, we build the large-scale dataset named ERF-X170FPS (High Resolution \textbf{E}vents and \textbf{R}GB \textbf{F}rames with e\textbf{X}treme Motions at \textbf{170FPS}) with the beam-splitter-based camera rigs comprising of the Prophesee Gen4 event cameras~\cite{internet_prohesee} (1280$\times$720) and Blackfly-S global shutter cameras (1440$\times$1080, maximum 226fps).
These two cameras are hardware-level synchronized using a micro-controller (the external trigger).
As summarized in Tab.~\ref{tab:datasets}, the proposed dataset has higher frame rates and larger spatial resolution enabling users to adjust the degrees of motion speeds and occlusions. 
Also, we capture more diverse scenes including deformable and fast-moving objects (dancer, animals, soccer, tennis, pop-corns, exploding coke, windmill, fountain, \etc.), static sceneries (building, landscape, flower, \etc) and driving scenarios with various distance ranges and diverse camera motion speeds compared to the previous arts~\cite{timelens++,wang2020joint}.
Please refer to \textit{suppl. materials} for a more detailed analysis.
\begin{figure*}[!t]
  \centering
  \includegraphics[width=1.9\columnwidth]{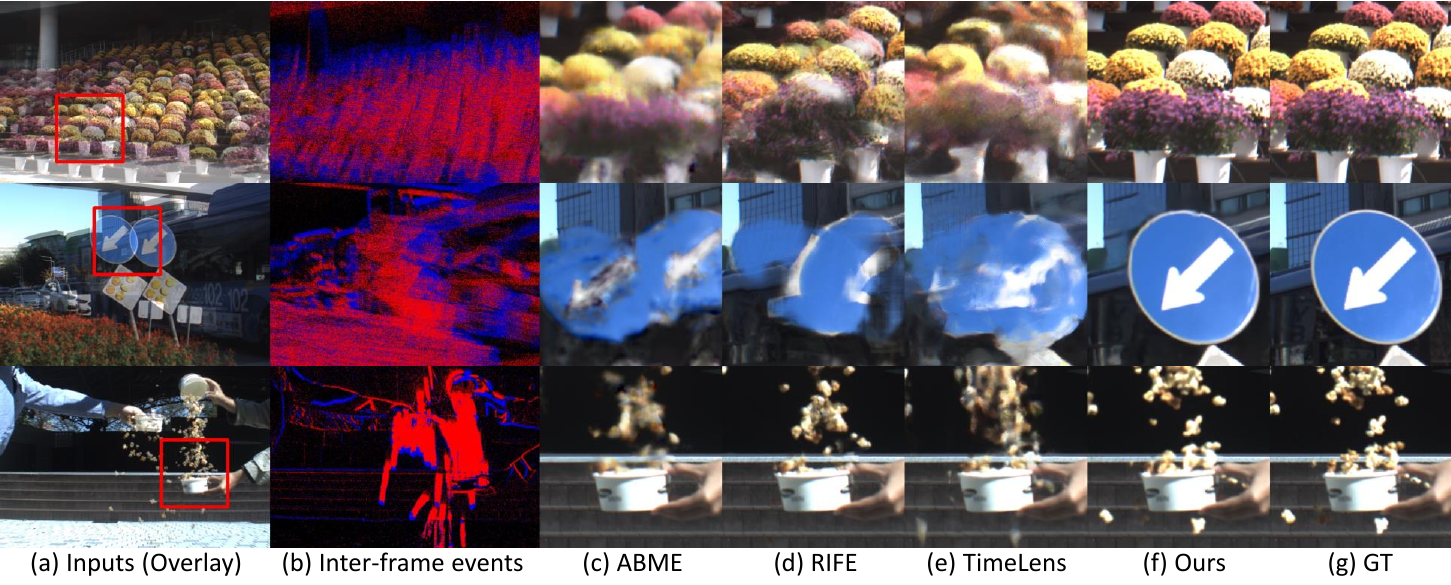}
  \vspace{-2mm}
  \caption{Visual results on the proposed ERF-X170FPS dataset. (Best viewed when zoomed in.)}\label{fig:qualitative_ERF}
\vspace{-3mm}
\end{figure*}

\section{Experiments}
\vspace{-0.5mm}
\subsection{Datasets and Implementation Details} 
\vspace{-1mm}
\noindent\textbf{Synthetic Event Datasets.} We train and test our framework on the GoPro~\cite{nah2017deep} dataset. 
For the additional comparison, we test the same model on the test split of Adobe240fps~\cite{su2017deep} dataset.
We generate the events between the consecutive video frames using the event simulator (ESIM)~\cite{Rebecq18corl}.
To validate the VFI performance according to the various motion ranges, we set the frame skip range to $\{7,15,23,31\}$ for the evaluation.
\\
\noindent\textbf{Real Event Datasets.} To verify the VFI performance on the real events, we conducted the experiments on the two publicly available datasets, High Quality Frame (HQF)~\cite{HQF_dataset} and BS-ERGB~\cite{timelens++}, and our ERF-X170FPS dataset.
\\
\noindent\textbf{Implementation Details.} We designed two models, ``Ours" and ``Ours-Large", that have same architecture but different parameters. Unless otherwise specified, our method afterwards are all ``Ours". More details are in \textit{suppl. material.}

\subsection{Experimental Results}
\vspace{-0.7mm}

\noindent\textbf{Synthetic Event Datasets.} Tab.~\ref{tab:synthetic} shows the quantitative comparison of our framework with the frame-based and event-based VFI methods on GoPro and Adobe240fps datasets.
Following the evaluation protocol of the previous event-based VFI methods~\cite{timelens,timereplayer,a2of}, we skip the video frames using original video sequences and compare the middle and whole skipped video frames with ground truth frames.
As in the previous works~\cite{timelens,timereplayer,a2of}, we directly evaluate SoTA frame-based VFI methods~\cite{superslowmo,DAIN,sepconv,BMBC,ABME,RIFE,IFRNet,vfiformer} using the official pretrained models.

As shown in Tab.~\ref{tab:synthetic}, our method outperforms all frame- and event-based methods in both two datasets with respect to all metrics.
Compared to the best frame-based method, ABME, our method shows significant performance gaps (\textbf{10.05}dB) in terms of PSNR when 7skips in the GoPro dataset.
As we maximize the frame skips of videos to 31, the performance gap between ABME and our method increases to \textbf{16.31}dB in the GoPro dataset.
Compared to the event-based VFI methods, our method still shows better results from \textbf{1.82}dB to \textbf{4.45}dB. 
\\
\noindent\textbf{Real Event Datasets.}
We compare our method and SoTA event-based VFI methods (TimeLens~\cite{timelens}, TimeLens++~\cite{timelens++}, TimeReplayer~\cite{timereplayer}, A$^2$OF~\cite{a2of}) on two publicly available real-event datasets, HQF~\cite{HQF_dataset} and BS-ERGB~\cite{timelens++} datasets.
As shown in Tab.~\ref{tab:real_events}, we record the highest PSNR and SSIM with significant margin compared to other methods in both datasets. 
In the HQF dataset, ours record the performance boost of \textbf{0.49}$\sim$\textbf{3.36}dB in 1skip and \textbf{0.89}$\sim$\textbf{3.92}dB in 3skips, using approximately \textbf{3.5} to \textbf{4.7} times fewer parameters than SoTA event-based VFI methods.
Moreover, the performance gaps between ``Ours-Large" and SoTA event-based VFI methods widen to \textbf{0.83}$\sim$\textbf{3.7}dB in 1skip and \textbf{1.23}$\sim$\textbf{4.26}dB in 3skips.
In the BS-ERGB dataset, our method also perform the best among other methods.
Ours-large model observes the PSNR improvement of \textbf{0.87}$\sim$\textbf{1.07}dB in 1skip and \textbf{0.96}$\sim$\textbf{1.01}dB in 3skips with \textbf{2.5}$\sim$\textbf{3.3} times less parameters.

\noindent\textbf{ERF-X170FPS Dataset.}
Finally, we perform the comparison on the splits of the proposed ERF-X170FPS dataset.
The test split of the dataset is composed of 36 different scenes containing significant non-linear motion and extreme scenes in high resolution.
For a fair comparison, we finetune the only public event-based VFI method~\cite{timelens} on our trainset with the same number of epochs as ours.

Tab.~\ref{tab:ERF} shows the quantitative results on the ERF-X170FPS dataset.
We can see that the proposed framework outperforms all SoTA VFI methods by a significant margin.
In particular, our framework outperforms the SoTA frame-based method up to \textbf{8.2}dB (PSNR) and \textbf{0.17} (SSIM).
Also, we show the significant performance boost of \textbf{7.9}dB (PSNR) and \textbf{0.159} (SSIM) compared to the SoTA event-based method~\cite{timelens}.
In the largest frame skips (11skips), the performance gap between ours and other methods becomes larger. Our method surpasses the best frame-based ABME and event-based TimeLens with \textbf{8.84/8.62}dB (PSNR) and \textbf{0.197/0.191} (SSIM), respectively.
Since the test split of the ERF-X170FPS dataset includes various non-linear and large camera motions with dynamic textures, it is challenging to interpolate the accurate intermediate frames.
Nevertheless, our framework can precisely estimate the inter-frame motion fields and interpolate the outputs.
Since TimeLens uses event-only motion fields and multi-stage image level-fusion methods, the error of each stage tends to accumulate recursively, leading to performance degradation.

Fig.~\ref{fig:qualitative_ERF} shows the visual results of ours and other SoTA methods. Our results in Fig.~\ref{fig:qualitative_ERF}(f) demonstrate that the proposed method produces precise interpolated frames even when the pixel displacement between input frames is vast and the motion is non-linear.
When other methods fail to interpolate accurate frames, our method successfully captures the vast and non-linear inter-frame motion
of tiny flowers, fast-moving traffic sign, and countless pop-corns.


\begin{table}[t]
\centering
\caption{Quantitative evaluation on publicly available real event datasets (BS-ERGB~\cite{timelens++} and HQF~\cite{HQF_dataset} datasets). We evaluate the performance based on whole frames between the input frames.}
\label{tab:real_events}
\vspace{-8pt}
\resizebox{\columnwidth}{!}{
\setlength{\tabcolsep}{1.5pt}
\begin{tabular}{c|cccc|cccc|c}
\hline
\multirow{3}{*}{Methods} & \multicolumn{4}{c|}{BS-ERGB~\cite{timelens++}} & \multicolumn{4}{c|}{HQF~\cite{HQF_dataset}} & \multirow{3}{*}{\begin{tabular}[c]{@{}c@{}}Params.\\ {[}M{]}\end{tabular}} \\ \cline{2-9}
 & \multicolumn{2}{c}{1skip} & \multicolumn{2}{c|}{3skips} & \multicolumn{2}{c}{1skip} & \multicolumn{2}{c|}{3skips} &  \\ \cline{2-9}
 & PSNR & SSIM & PSNR & SSIM & PSNR & SSIM & PSNR & SSIM &  \\ \hline
TimeLens$^{\dagger}$~\cite{timelens} & 28.36 & - & 27.58 & - & 32.49 & 0.927 & 30.57 & 0.900 & 72.2 \\
TimeReplayer$^{\dagger}$~\cite{timereplayer} & - & - & - & - & 31.07 & 0.931 & 28.82 & 0.866 & - \\
TimeLens++$^{\dagger}$~\cite{timelens++} & 28.56 & - & 27.63 & - & - & - & - & - & 53.4 \\ 
A$^2$OF$^{\dagger}$~\cite{a2of} & - & - & - & - & 33.94 & 0.945 & 31.85 & 0.932 & - \\ \hline
Ours & \underline{29.32} & \underline{0.815} & \underline{28.46} & \underline{0.806} & \underline{34.43} & \underline{0.949} & \underline{32.74} & \underline{0.934} & 15.4 \\
Ours-Large & \textbf{29.43} & \textbf{0.816} & \textbf{28.59} & \textbf{0.808} & \textbf{34.77} & \textbf{0.953} & \textbf{33.08}  & \textbf{0.940} & 22.2 \\ \hline
\end{tabular}
}
 \vspace{-3mm}
\end{table}

\begin{table}[t]
\centering
\caption{Quantitative evaluation based on middle frame on the ERF-X170FPS dataset with state-of-the-art VFI methods.}
\label{tab:ERF}
\vspace{-8pt}
\resizebox{\columnwidth}{!}{
\setlength{\tabcolsep}{2pt}
\begin{tabular}{c|cccccccc}
\hline
\multirow{3}{*}{Methods} & \multicolumn{8}{c}{ERF-X170FPS} \\ \cline{2-9} 
 & \multicolumn{2}{c}{3skips} & \multicolumn{2}{c}{7skips} & \multicolumn{2}{c|}{11skips} & \multicolumn{2}{c}{Avg.} \\ \cline{2-9} 
 & PSNR & SSIM & ~~PSNR & SSIM & ~~PSNR & \multicolumn{1}{c|}{SSIM} & PSNR & SSIM \\ \hline
ABME~\cite{ABME} & 27.78 & 0.861 & ~~22.68 & 0.748 & ~~19.96 & \multicolumn{1}{c|}{0.679} & 23.47 & 0.763  \\
VFIformer~\cite{vfiformer} & 26.29 & 0.836 & ~~21.38 & 0.724 & ~~19.04 & \multicolumn{1}{c|}{0.661} & 22.24 & 0.740 \\
IFRNet~\cite{IFRNet} & 26.03 & 0.826 & ~~21.43 & 0.712 & ~~19.14 &  \multicolumn{1}{c|}{0.648} & 22.20 & 0.729 \\
RIFE~\cite{RIFE} & 26.99 & 0.841 & ~~22.18 & 0.723 & ~~19.64 & \multicolumn{1}{c|}{0.652} & 22.94 & 0.739 \\ 
TimeLens$^{\dagger}$~\cite{timelens} & 25.34 & 0.807 & ~~21.99 & 0.729 & ~~20.18 & \multicolumn{1}{c|}{0.685} & 22.50 & 0.740 \\ \hline 
Ours & \underline{32.29} & \underline{0.924} & ~~\underline{30.12} & \underline{0.897} & ~~\underline{28.80} & \multicolumn{1}{c|}{\underline{0.876}} & \underline{30.40} & \underline{0.899} \\ 
Ours-Large & \textbf{32.37} & \textbf{0.926}  & ~~\textbf{30.22} & \textbf{0.899} &  ~~\textbf{28.94} & \multicolumn{1}{c|}{\textbf{0.878}} & \textbf{30.51} & \textbf{0.901} \\ \hline
\end{tabular}
}
\vspace{-3mm}
\end{table}

\vspace{-5pt}
\subsection{Model Analysis}
\vspace{-2pt}
\noindent\textbf{EIF-BiOFNet.}
We compare the inter-frame motion fields of ours with frame-based~\cite{BMBC,superslowmo,ABME,RIFE} and event-based~\cite{timelens} methods in Tab.~\ref{tab:inter_frame_motion_fields}.
For a fair comparison, we retrained the SoTA event-based inter-frame motion fields module~\cite{timelens} with the same trainset as ours.
We perform backward warping of input frames using the estimated motion fields and calculate the PSNRs of the warped frames.
RIFE shows the best performance among the frame-based methods and asymmetrical flow-based method (ABME) shows better performance than linear approximation-based methods (BMBC, SuperSloMo).
The SoTA event-based method, TimeLens-Flow~\cite{timelens}, shows higher PSNRs than all the frame-based methods utilizing the abundant temporal information of the events.
Nevertheless, ours outperforms TimeLens-Flow significantly by avg. 2.58dB and 3.76dB in 31skips using 2.6 times fewer parameters. It demonstrates that our EIF-BIOFNet modules play the critical role for accurate inter-frame motion field estimation.

In Tab.~\ref{ablation_motion_fields}, we report the performance contribution of each component in EIF-BiOFNet. Using only E-BiOF shows a similar performance to TimeLens-Flow and using only I-BiOF shows higher avg. PSNRs by 1.08dB compared to E-BiOF. When all three modules are used, we observe a significant performance improvement of up to 2.58dB. It means that each module of EIF-BiOFNet elaborately utilizes and fuses cross-modal information.
\\
\noindent\textbf{Frame Synthesis Network.}
We investigate the influence of each component in the interactive attention-based frame synthesis network.
We first analyze the warping and synthesis-based interpolation.
For the ablation, we replace the transformer decoder layer with a CNN-based decoder with a similar number of parameters.
We observe a performance boost (+1.04dB) when using the warping-based feature using inter-frame motion fields obtained from our EIF-BiOFNet instead of synthesis-based features.
When we fuse the synthesis and warping-based features, performance increases by 1.86dB compared with the baseline.
With the transformer decoder layer, 1.43dB PSNR inclines by replacing the CNN-based decoder(the decoder part of UNet) with the TIA (Transformer Interactive Attention) blocks as in the 3rd and 5th rows of Tab.~\ref{ablation_tia}.
We also confirm the performance increases by 0.13dB with the TSA (Transformer Self-Attention) layers for further refining output as in the 5th and 6th rows.
The results demonstrate that TIA blocks better combine the warping- and synthesis-based features.

\begin{table}[t]
\centering
\caption{Quantitative comparison with state-of-the-art inter-frame motion field estimation methods on GoPro dataset.}
\label{tab:inter_frame_motion_fields}
\vspace{-8pt}
\resizebox{\columnwidth}{!}{
\setlength{\tabcolsep}{3pt}
\begin{tabular}{c|cccc|c|c}
\hline
\multirow{2}{*}{Methods} & 7skips & 15skips & 23skips & 31skips & Avg. & \multirow{2}{*}{\begin{tabular}[c]{@{}c@{}}Params.\\ {[}M{]}\end{tabular}} \\ \cline{2-6}
 & PSNR & PSNR & PSNR & PSNR & PSNR &  \\ \hline
SuperSloMo~\cite{superslowmo} & 26.22 & 21.98 & 20.04 & 18.91 & 21.79 & 19.79 \\
BMBC~\cite{BMBC} & 26.53 &  21.51 & 19.29 & 18.09 & 21.36 & 10.56 \\
ABME~\cite{ABME} & 26.30 & 22.35 & 20.42 & 19.34 & 22.10 & 11.55 \\ 
RIFE~\cite{RIFE} & 26.82 & 22.40 & 20.40 & 19.31 & 22.23 & 3.04 \\ 
TimeLens-Flow$^{\dagger}$~\cite{timelens} & \underline{30.79} & \underline{28.64} & \underline{26.75} & \underline{25.18} & \underline{27.84} & 19.81  \\ \hline
Ours-Flow & \textbf{32.29} & \textbf{30.73} & \textbf{29.71} & \textbf{28.94} & \textbf{30.42} & 7.64 \\ \hline
\end{tabular}
}
\vspace{-3mm}
\end{table}

\begin{table}[t]
\centering
\caption{Ablation study of the EIF-BiOFNet on GoPro dataset.}
\label{ablation_motion_fields}
\vspace{-8pt}
\resizebox{\columnwidth}{!}{
\setlength{\tabcolsep}{4pt}
\begin{tabular}{ccccccc|c}
\hline
\multicolumn{3}{c}{Components} & 7skips & 15skips & 23skips & 31skips & Avg. \\  \hline
E-BiOF & I-BiOF & F-BiOF & PSNR & PSNR & PSNR & PSNR & PSNR \\ \hline
\checkmark &  &  & \underline{30.72} & 28.66 & 26.77 & 25.21 & 27.84 \\
 & \checkmark &  & 30.27 & \underline{29.28} & \underline{28.43} & \underline{27.69} & \underline{28.92} \\
\checkmark & \checkmark & \checkmark & \textbf{32.29} & \textbf{30.73} & \textbf{29.71} & \textbf{28.94} & \textbf{30.42}  \\ \hline
\end{tabular}
}
\vspace{-2mm}
\end{table}

\begin{table}[t]
\centering
\caption{Ablation study of the interactive attention-based frame synthesis network on GoPro dataset.}
\label{ablation_tia}
\vspace{-8pt}
\resizebox{\columnwidth}{!}{
\setlength{\tabcolsep}{4pt}
\begin{tabular}{cccccccc|c}
\hline
\multicolumn{4}{c}{Components} & 7skips & 15skips & 23skips & 31skips & Avg. \\ \hline
Syn. & Warp & TSA & TIA & PSNR & PSNR & PSNR & PSNR & PSNR \\ \hline
\checkmark &  &  &  & 34.77 & 33.53 & 32.69 & 31.71 & 33.18 \\
 & \checkmark &  &  & 35.55 &  34.53 & 33.81 & 32.97 & 34.22 \\
\checkmark & \checkmark &  &  & 36.06 & 35.23 & 34.74 & 34.11 & 35.04 \\
\checkmark & \checkmark & \checkmark & & 37.30 & 36.25 & 35.71 & 35.09 & 36.09 \\
\checkmark & \checkmark & & \checkmark & \underline{37.71} & \underline{36.62} & \underline{36.07} & \underline{35.49} & \underline{36.47} \\
\checkmark & \checkmark & \checkmark & \checkmark & \textbf{37.84} & \textbf{36.74} & \textbf{36.19} & \textbf{35.62} & \textbf{36.60} \\ \hline
\end{tabular}
}
\vspace{-6mm}
\end{table}

\vspace{-5pt}
\section{Conclusion}
\vspace{-4pt}
We propose an event-based VFI method with asymmetric inter-frame motion fields by elaborately handling the characteristics of the modality of the events and frame. 
Extensive experiments demonstrate that the proposed methods show remarkable improvement over the SoTA VFI methods.
We also construct the ERF-X170FPS dataset with HFR frame-based and HR event cameras, which will be valuable for the event-based VFI research community. \\
\small{\textbf{Acknowledgment}} This work was supported by the National Research Foundation of Korea(NRF) grant funded by the Korea government(MSIT)(NRF2022R1A2B5B03002636).

{\small
\bibliographystyle{ieee_fullname}
\bibliography{egbib}
}

\clearpage

\appendix

\noindent {\LARGE \textbf{Supplementary Materials}} \\[5pt]

\bigskip 

\vspace{-5mm} 






\section{ERF-X170FPS Dataset}
\subsection{Camera Setup Details}
\begin{figure}[t]
\centering
\includegraphics[width=0.85\linewidth]{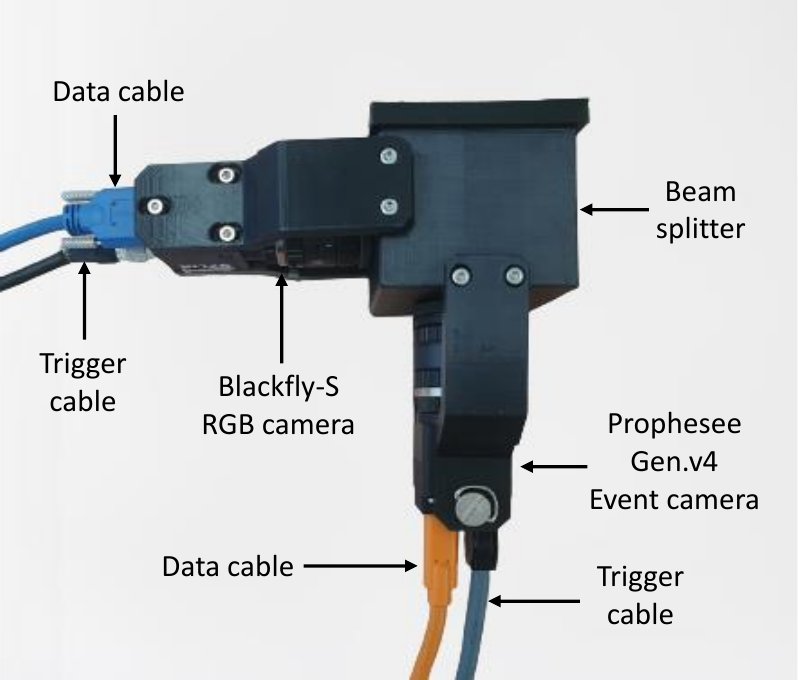} 
\vspace{-1mm}
\caption{Our beam-splitter-based camera setup.}
\label{supp:beam_splitter_camera_setup}
\vspace{-4mm}
\end{figure}

\begin{figure*}[!b]
\centering
\includegraphics[width=0.95\linewidth]{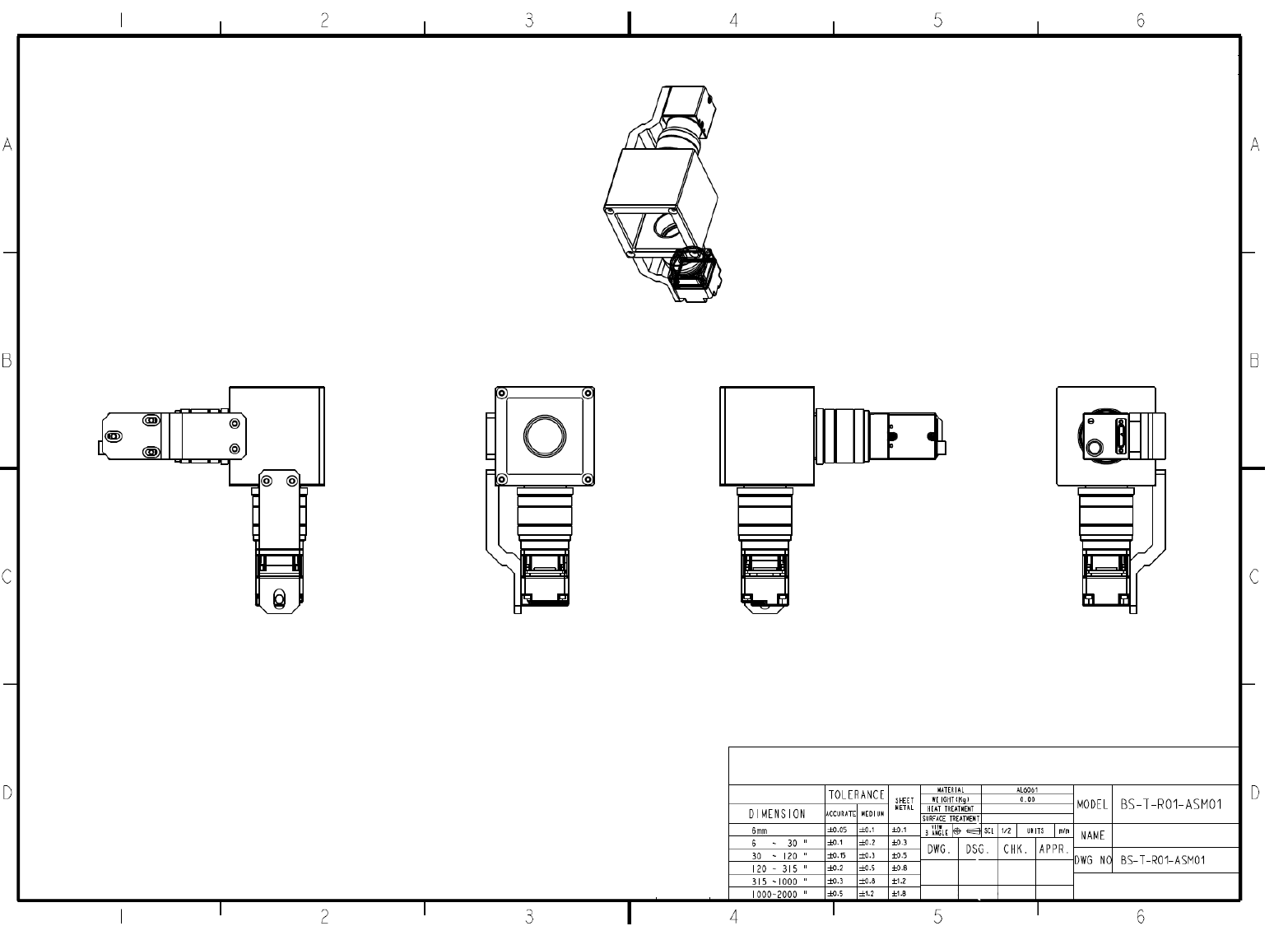} 
\caption{Beam-splitter-based camera rig 3D CAD drawing.}
\label{supp:beam_splitter_3dcad}
\end{figure*}

\noindent \textbf{Beam-splitter-based camera setup} \ 
A beamsplitter is an optical device for splitting incident light into two beams according to a specified ratio.
Therefore, a beamsplitter enables two different cameras to capture the same scenes by the split light source.
The beam-splitter-based camera setup for photographing the ERF-X170FPS dataset is shown in Fig.\ref{supp:beam_splitter_camera_setup}.
For beam-splitter selection, we choose a non-polarized cube beam-splitter rather than the plate-based beam-splitter to alleviate beam-shifting issues. 
We select \textit{BS-CUBE-NON-POL-VIS-50MM-TS} beam-splitter with a size of $50cm^3$, which can capture a large field of view of the scenes.
The beamsplitter splits the incident light into non-polarized light in a ratio of 50:50.
After that, we designed a 3D-CAD model for a rigid camera rig that can completely immobilize two cameras and a beam-splitter, as shown in Fig.\ref{supp:beam_splitter_3dcad}.
For the RGB camera, we select \textit{FLIR BFS-U3-16S2C-CS}.
The camera can shoot videos at the resolution of $1440\times1080$ and up to 226FPS and support an external trigger interface.
Also, we selected \textit{EVK4 HD Prohesee Gen4.1 HD} event camera. 
The event camera can capture videos with a resolution of $1280\times720$.
We then fixed these two cameras and a beam-splitter to the designed camera rig.
As a result, two cameras can receive the incoming co-axis light source at a fixed position.
\\
\textbf{Camera synchronization} \
\textit{In practice, we can't obtain the accurate timestamps of two cameras without interfacing with the external trigger.}
For this reason, we designed a micro-controller(ATmega328) as an external trigger for hard-ware level synchronization of the event and RGB camera.
The event and RGB cameras are connected to the microcontroller through a trigger cable, as shown in Fig.\ref{supp:beam_splitter_camera_setup}.
Therefore, the generated signals of the microcontroller are simultaneously transmitted to the event camera and RGB camera, respectively.
After that, we create recording software using provided C++ SDK of each camera product to control these two cameras by receiving signals from the microcontroller.
Each camera receives the falling edge and the rising edge of trigger signals and performs synchronization with the period of the signal.
Through this external trigger, we can control the RGB camera's frame rate and exposure time with synchronized signals.
Also, we obtain accurate timestamps of the events between two consecutive standard frames.
As a result, we can obtain two different modality data with precise timestamp information as the two cameras are synchronized at the hardware level. \\
\textbf{Calibration} \ 
Two cameras receive a co-axis light source due to beam-splitter camera setup.
As a result, two cameras have the minimal baselines.
However, they have different fields of view due to the different sensor sizes of each camera.
To this end, we calibrate the event and RGB camera for spatially aligning two different modality data.
For intrinsic and extrinsic calibration, we use a blinking checkerboard pattern.
After the calibration process, we transform the spatial pixel position of the events using the estimated homography matrix.
We then crop the standard frames whose fields of view do not overlap with the event camera.
As a result, we simultaneously record spatially aligned event and frame data with a resolution of $1440\times975$.

\begin{table*}[!t]
\centering
\captionsetup{font=small}
\caption{The overview of test set of ERF-X170FPS.}
\label{tab:ERF_X170FPS_test_details}
\vspace{-2mm}
\scalebox{0.81}{
\begin{tabular}{cccc}
\hline
Seq.Name & Camera settings & Explanations & Scene class \\ \hline
Building 01 & 170FPS, 990 frames & capturing with fast camera movement of building. &  (ii) \\
Traffic load 01 & 170FPS, 990 frames & capturing fast car and bicycles with zig-zag camera motion.  & (ii) \\
Fountain water pump 01 & 170FPS, 990 frames &  capturing fountain with non-linear camera motion. & (ii) \\
Fountain water pump 02 & 170FPS, 990 frames & capturing fountain with up-down camera motion. & (ii) \\
Flowers 01 & 170FPS, 990 frames & capturing flowers with rotations. & (ii) \\
Geese and lake 01 & 170FPS, 990 frames & capturing with moving geese in the lake. & (ii) \\
Traffic road 01 & 170FPS, 990 frames &  capturing fast moving cars and traffic signs.  & (ii)  \\
Dancer 01 & 170FPS, 990 frames  &  capturing fast moving dancer with non-linear camera motion. & (iii) \\
Dancer 02 & 170FPS, 990 frames & capturing dancer with close distance. & (iii)  \\
Geese swarm 01 & 170FPS, 990 frames & capturing geese swarm.  & (iii)  \\
Windmill 01 & 170FPS, 990 frames & capturing fast rotating windmill. & (iii) \\
Bicycle road 01 & 170FPS, 990 frames & capturing bicycle road with fast up-down camera movement. & (ii) \\
Traffic road 02 & 170FPS, 990 frames & capturing fast moving car with fast camera movement. & (ii) \\
Soccer players 01 & 170FPS, 990 frames & capturing fast soccer players. & (iii) \\
Soccer players 02 & 170FPS, 990 frames & capturing fast soccer players. & (iii) \\
Soccer players 03 & 170FPS, 990 frames & capturing fast dribbling soccer players. & (iii) \\
Driving forest 01 & 170FPS, 990 frames & capturing the forest on the car. & (i) \\
Driving u-turn 01 & 170FPS, 990 frames & capturing trees with car u-turn. & (i) \\
Driving forest 02 & 170FPS, 990 frames & capturing the forest on the car. & (i) \\
Driving bicycle stand 01 & 170FPS, 990 frames & capturing the bicycle stands on fast car. & (i) \\
Tennis 01 & 170FPS, 990 frames & capturing tennis players with irregular left-right camera motion. &  (iii) \\
Tennis 02 & 170FPS, 990 frames & capturing tennis players. &  (iii) \\
Driving u-turn 02 & 170FPS, 990 frames & capturing the scenes on the fast u-turning car. & (i) \\
Driving urban 01 & 170FPS, 990 frames & capturing the urban scenes on the fast moving car. & (i) \\
Driving left-turn 01 & 170FPS, 990 frames & capturing with a left-turn on an intersection and shooting cars and trees. & (i) \\
Driving bridge 01 & 170FPS, 990 frames & capturing the bridge on the fast moving car. & (i) \\
Driving department store 01 & 170FPS, 990 frames & capturing the department store on the fast moving car. & (i) \\
Driving road 01 & 170FPS, 990 frames & capturing the trees and national police agency on the fast moving car. & (i) \\
Falling pop-corn 01 & 170FPS, 990 frames & capturing falling pop-corn. & (iv) \\
Climbing people 01 & 170FPS, 990 frames & capturing person quickly climbing stairs. & (iii) \\
Fast rotating man 01 & 170FPS, 990 frames & capturing fast rotating man. & (iii) \\
Exploding cola 01 & 170FPS, 990 frames & capturing exploding cola with mentos. & (iv) \\
Exploding cola 02 & 170FPS, 990 frames & capturing exploding cola with mentos. & (iv) \\
lakelet 01 & 170FPS, 990 frames & capturing lakelet with non-linear camera motion. & (iv) \\
Waterfall 01 & 170FPS, 990 frames & capturing water fall in the lakelet. & (iv) \\
Handwash 01 & 170FPS, 990 frames & capturing a person washing his hands. & (iv) \\
\hline
\end{tabular}
}
\end{table*}

\subsection{Photographing Dataset}
To properly evaluate VFI performance in diverse circumstances, it is essential to shoot various scenes, such as multiple camera motions and objects.
Specifically, in event-based VFI, scenes for synthesis-based interpolation and warping-based interpolation should be distributed harmoniously.
As mentioned in the main paper, synthesis-based interpolation is effective in regions where motion fields are invalid.
The synthesis-based interpolation works well in situations such as flooding water, rotating objects, fire, and occlusion of scenes.
However, in many typical situations, there are many areas where the motion fields are valid due to camera or object movement, except for the above cases.
In the case of the previous event-based VFI datasets~\cite{timelens++}, the test scenes of the BS-ERGB dataset mainly photographed the scenes where motion fields are invalid. (\eg, flooding water tank, fire, popping eggs, fast rotating objects, thin objects \textbf{with static camera movement}).
For these reasons, it is hard to evaluate the performance of motion-based frame interpolation methods in the previous event-based VFI dataset.
To alleviate the deficiency, we photographed our ERF-X170FPS dataset with the following multiple categories:
\\
\textbf{(i)} We capture moving objects on a fast-moving car with diverse vehicle speeds.  \\
\textbf{(ii)} We move the camera with irregular directions and speed to shoot static scenes and moving objects. (\eg, flowers, lake, fountain, road, windmill, traffic signs, building, animals, crowds, etc.)  \\
\textbf{(iii)} We photographed the situation of the dynamic motion of people and animals with fast camera movement.(\eg, dancers, soccer, tennis player, rotating men, etc)    \\
\textbf{(iv)} We photographed the fast deformable objects(\eg, water, exploding cola, rapidly falling and rotating objects, etc.). \\
In the case of (i), (ii), warping-based interpolation mainly works, and synthesis-based interpolation generally works well in (iv).
In the case of (iii), the interpolation result is the sum of the combination of two methods.
Based on this analysis, we photograph four possible situations in balanced numbers.
\subsection{ERF-X170FPS-Split}
\begin{figure*}[!b]
\centering
\includegraphics[width=0.95\linewidth]{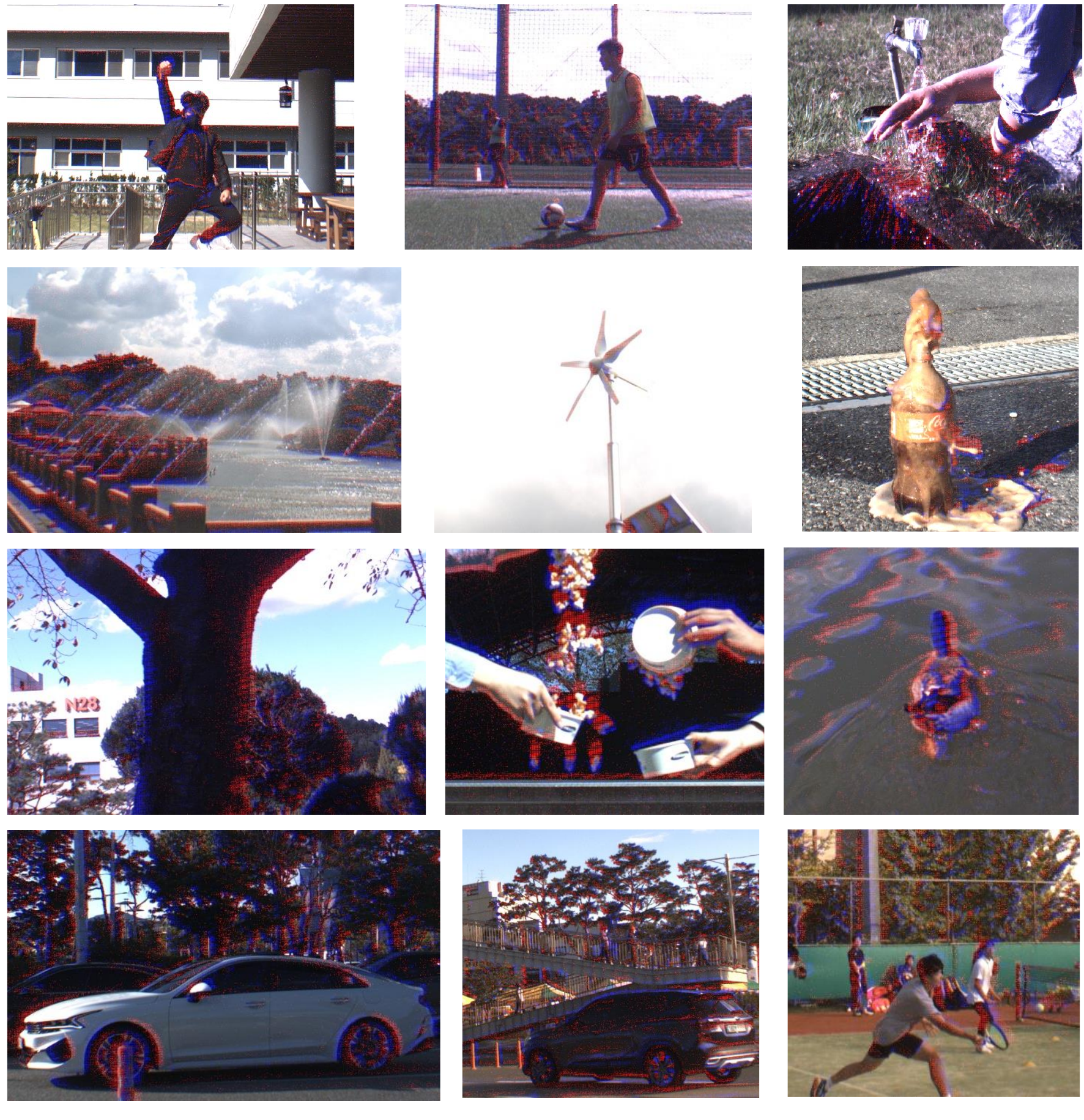} 
\caption{The examples of our ERF-X170FPS test dataset. Our dataset contains diverse scenes and motion speed at 170FPS, which consists of HR video frames and temporally synchronized HR event data.}
\label{supp:ERF_examples}
\end{figure*}

We manually selected 36 scenes for the test set of ERF-X170FPS in consideration of the degree of occlusion and motion speed.
As in the Tab.\ref{tab:ERF_X170FPS_test_details}, we have balanced the distribution of the above four situations.
Compared to the BS-ERGB~\cite{timelens++} dataset, the proposed ERF-X170FPS dataset is well distributed in four categories of situations to allow better evaluation of motion-based VFI methods. 
As mentioned in the main paper, we evaluated the $\{3,7,11\}$ skips of original videos to compare the other VFI methods for diverse motion ranges. 
The examples of our test set of ERF-X170FPS are shown in Fig.\ref{supp:ERF_examples}.
We divided the remaining scenes into validation and train sets.


\begin{table*}[!t]
\centering
\captionsetup{font=small}
\caption{Quantitative evaluation of multi-frame interpolation (whole skipped frames of 7skips) on the GoPro~\cite{nah2017deep} dataset.}
\label{tab:multi_frame_interp}
\vspace{-8pt}
\scalebox{0.8}{
\begin{tabular}{c|ccccccccc}
\hline
\multirow{2}{*}{Methods} & \multicolumn{9}{c}{GoPro} \\ \cline{2-10} 
 & SuperSloMo~\cite{superslowmo} & SepConv~\cite{sepconv} & DAIN~\cite{DAIN} & \multicolumn{1}{c|}{BMBC~\cite{BMBC}} & TimeLens$^{\dagger}$~\cite{timelens} & TimeReplayer$^{\dagger}$~\cite{timereplayer} & \multicolumn{1}{c|}{A$^2$OF$^{\dagger}$~\cite{a2of}} & Ours & Ours-Large \\ \hline
PSNR & 28.95 & 29.13 & 28.81 & \multicolumn{1}{c|}{29.08} & 34.81 & 34.02 & \multicolumn{1}{c|}{36.61} & \underline{37.77} & \textbf{38.03} \\
SSIM & 0.876 & 0.876 & 0.876 & \multicolumn{1}{c|}{0.875} & 0.959 & 0.960 & \multicolumn{1}{c|}{0.971} & \underline{0.974}  & \textbf{0.975} \\ \hline
\end{tabular}
}
\end{table*}
\section{Additional Experimental Results and Details}
\label{sec:experimtents}
\subsection{Video Demos}
We generated demo videos on the proposed ERF-X170FPS and GoPro/HQF datasets. 
Demo videos named as \emph{\textcolor{red}{$\mathrm{Video\_demo.mp4}$}} include the qualitative comparison of videos with other SoTA VFI methods.

\subsection{Quantitative evaluation results of multi-frame interpolation on GoPro~\cite{nah2017deep} dataset}
In the main paper, we report the evaluation results for the middle frame of the skipped video frames on the synthetic event datasets in the Tab.~\ref{tab:multi_frame_interp}.
We additionally perform comparison on the whole frames of the skipped video frames (7skips in GoPro datasets).
As with the main paper, we significantly outperform frame-based and event-based video frame interpolation methods.

\subsection{Datasets Details}
\noindent\textbf{Real Event Datasets} As mentioned in the main paper, we conducted in two publicly available real-event datasets.
The first dataset is High Quality Frame (HQF)~\cite{HQF_dataset} dataset captured by the DAVIS-240C event camera with 14 different scenes.
This dataset provides the synchronized events and frames (240$\times$180 resolution) with 14 different scenes.
In addition, we conduct the experiments on the BS-ERGB~\cite{timelens++} dataset.
Following the evaluation protocol with the previous methods~\cite{timelens,timelens++,timereplayer,a2of}, we evaluate whole skipped frames within $\{1,3\}$ frame skips for both datasets.

\subsection{Implementation Details}
We implemented our framework using PyTorch~\cite{paszke2019pytorch}. To train our networks, we use batch size 6 for the all datasets and AdamW~\cite{loshchilov2017decoupled} optimizer to update network weight using initial learning rate $1e^{-4}$ and decay rate 0.5.
We apply random cropping to the frame and events for the same pixel position.
For the quantitative evaluation, we use the standard evaluation metrics, PSNR and SSIM~\cite{wang2004image}.


\subsection{Qualitative comparison of the warped frame on GoPro~\cite{nah2017deep} dataset}
\begin{figure*}[!t]
\centering
\vspace{-3mm}
\includegraphics[width=1.0\linewidth]{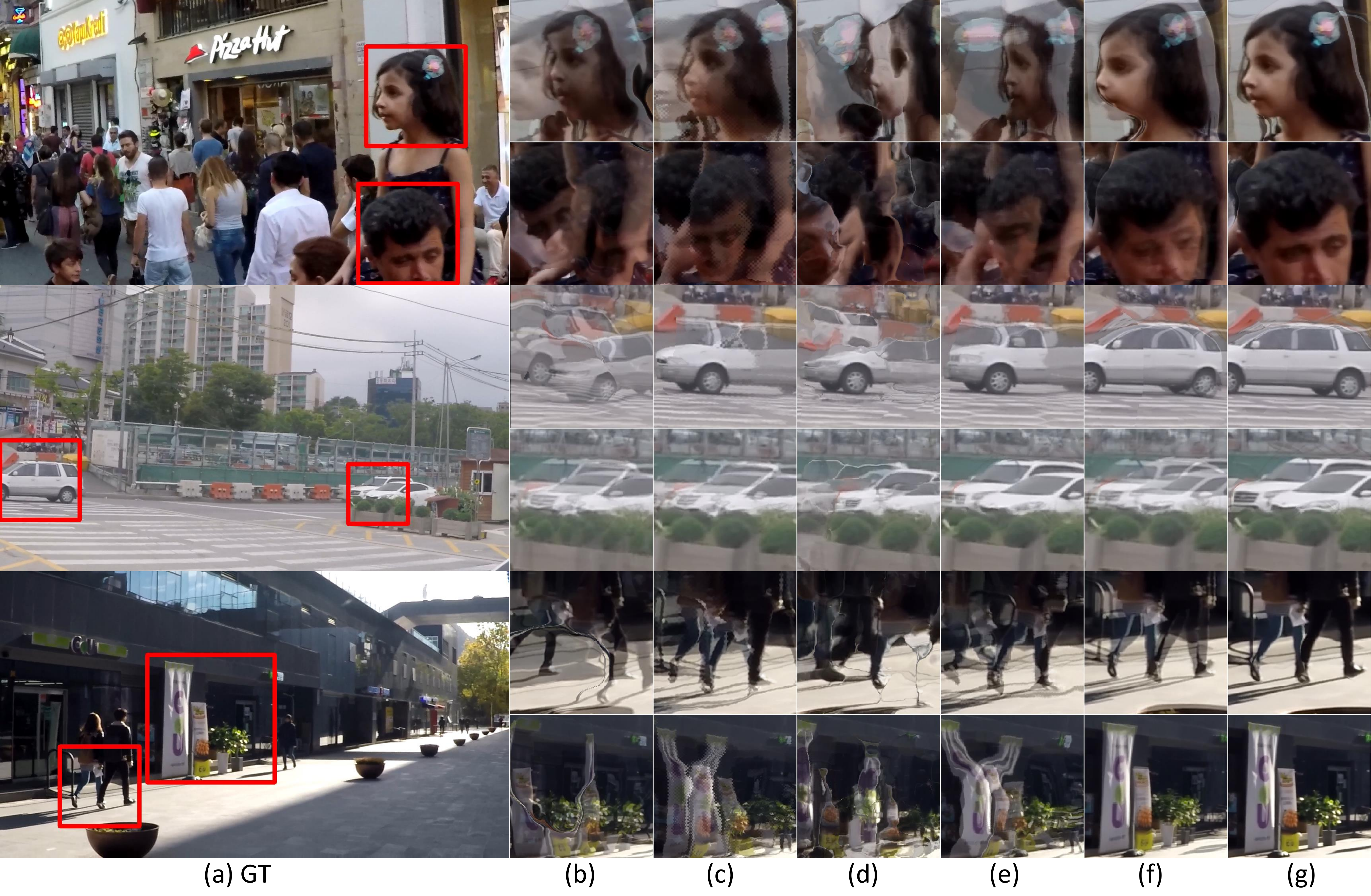} 
\vspace{-5mm}
\caption{The qualitative comparison on the warped frame of estimated inter-frame motion fields on GoPro dataset. In order, (a) GT frame, (b) SuperSloMo~\cite{superslowmo} (c) BMBC~\cite{BMBC} (d) ABME~\cite{ABME} (e) RIFE~\cite{RIFE} (f) TimeLens~\cite{timelens} (g) Ours. As in the results, we confirm that our method produces more accurate warped frames than state-of-the-art inter-frame motion fields estimation methods. \textbf{Please zoom for better visualization.}}
\label{supp:warped_res_gopro}
\end{figure*}

Due to the lack of the main paper, we only report the quantitative results of the warped frame of inter-frame motion fields in the main paper.
In addition to the quantitative results, we perform qualitative comparison of  estimated inter-frame motion fields in the Fig.\ref{supp:warped_res_gopro}
As shown in the figure, our EIF-BiOFNet more reliably estimate bidirectional inter-frame motion fields than state-of-the-art inter-frame motion field estimation methods~\cite{superslowmo,BMBC,RIFE,ABME,timelens}.

\subsection{How does EIF-BiOFNet operate w/o events?}
If there is no motion (with no events), the anchor and the boundary frames are the same, and the OF comes out as nearly zero.
The other case is that relative motion exists, but events are unavailable  (anchor and boundary frames differ).
For the second case, we could only train the I-BiOFNet, and the ablation results are shown in Tab.~\ref{tab:ablation}.
We can see that the anchor feature synthesis may not perform well compared to using events.
However, even without events, ours shows comparable performance to that of image-based VFI SoTA method ABME~\cite{ABME}

\subsection{Implementation details of ~\cite{timelens}}
For finetuning ~\cite{timelens} on the ERF-X170FPS datasets, we followed the original paper's approach by training each stage individually and adhering to the original training strategy.
Tab.~\ref{tab:timeLens_eval} demonstrates that the finetuned model outperforms the official pretrained model.

In the case of TimeLens-Flow, the pretrained model is trained on not only synthetic event datasets but also real-world event datasets. Therefore, the performance will be degraded if we directly apply the pre-trained model to the synthetic datasets. For a fair comparison with ours, we retrained the TimeLens-flow only on the GoPro dataset.

\begin{table}[!t]
\centering
\captionsetup{font=small}
\captionsetup{aboveskip=2pt, belowskip=0pt}
\caption{Ablation study of inter-frame motion fields without events on the GoPro datasets.}~\label{tab:ablation}
\scalebox{0.7}{
\begin{tabular}{c|c|cclc}
\hline
 & ABME {[}\textcolor{green}{30}{]} & I-BiOF (w/o ev.) & I-BiOF (w/ ev.) & E-BiOF & EIF-BiOF \\ \hline
PSNR & 22.1 & 21.9 & 28.9 & \multicolumn{1}{c}{27.8}& 30.4 \\ \hline
\end{tabular}
}
\end{table}

\begin{table}[!t]
\centering
\captionsetup{font=small}
\captionsetup{aboveskip=2pt, belowskip=0pt}
\caption{Quantitative evaluation results of TimeLens~\cite{timelens} on the ERF-X170FPS dataset. ~\cite{timelens}-P and ~\cite{timelens}-F represent the pretrained and finetuned model of TimeLens~\cite{timelens}, respectively.}~\label{tab:timeLens_eval}
\resizebox{\columnwidth}{!}{
\begin{tabular}{c|cccccc|cc}
\hline
\multirow{2}{*}{} & \multicolumn{2}{c}{3skips} & \multicolumn{2}{c}{7skips} & \multicolumn{2}{c|}{11skips} & \multicolumn{2}{c}{Avg.} \\ \cline{2-9} 
 & PSNR & SSIM & PSNR & SSIM & PSNR & SSIM & PSNR & SSIM \\ \hline
\cite{timelens}-P & \underline{23.08} & \underline{0.724} & \underline{20.76} & \underline{0.666} & \underline{19.44} & \underline{0.633} & \underline{21.09} & \underline{0.674} \\
\cite{timelens}-F & \textbf{25.34} & \textbf{0.807} & \textbf{21.99} & \textbf{0.729} & \textbf{20.18} & \textbf{0.685} & \textbf{22.50} & \textbf{0.740} \\ \hline
\end{tabular}}
\centering
\end{table}

\subsection{Additional Visual Results}

\subsubsection{More Visual Results on ERF-X170FPS dataset} 
\begin{figure*}[!t]
\centering
\vspace{-3mm}
\includegraphics[width=0.84\linewidth]{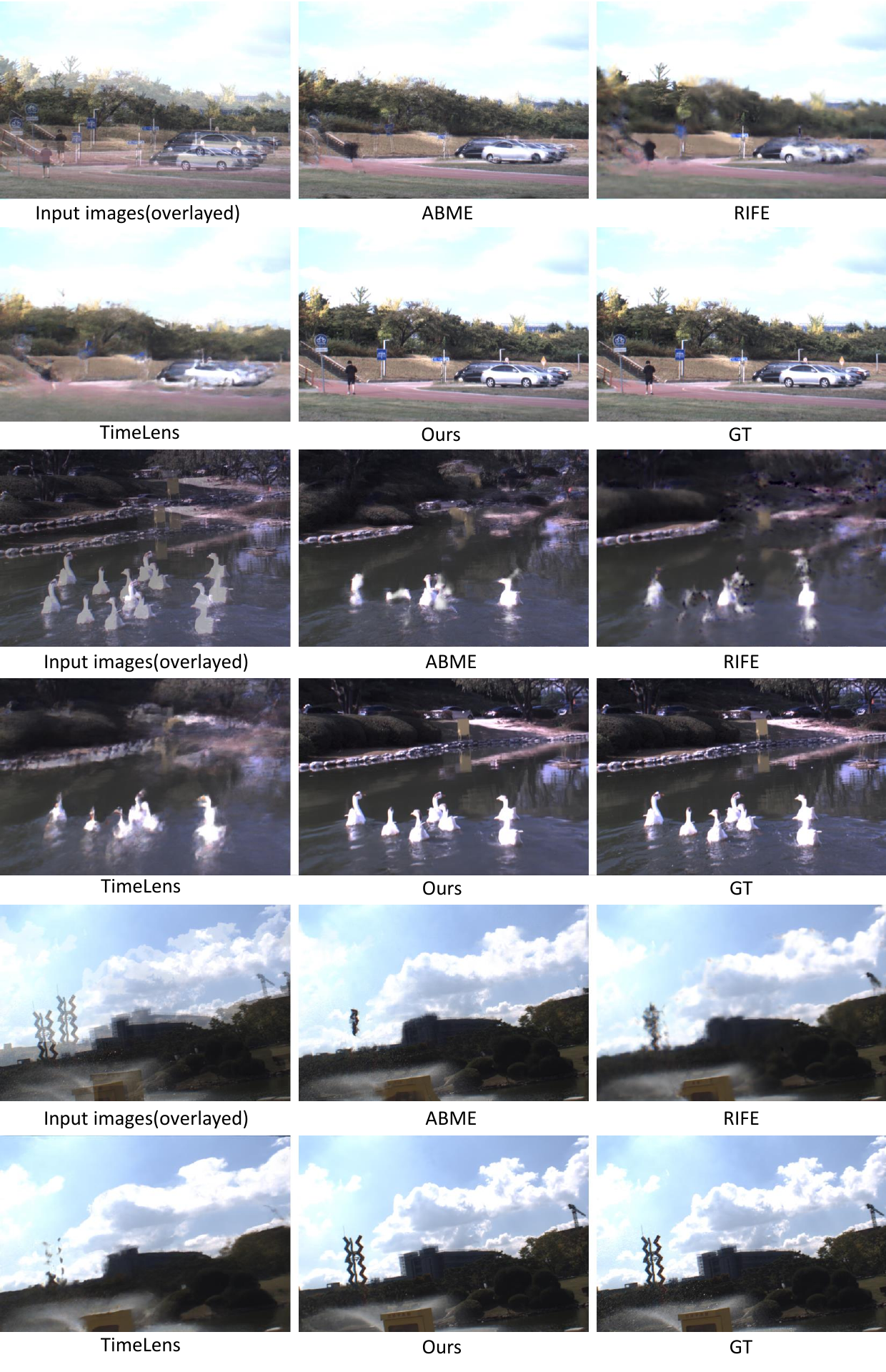} 
\vspace{-3mm}
\caption{Visual results on the ERF-X170FPS dataset. (Best viewed when zoomed in.)}
\label{supp:visual_res_erf1}
\end{figure*}

\begin{figure*}[!t]
\centering
\vspace{-3mm}
\includegraphics[width=0.84\linewidth]{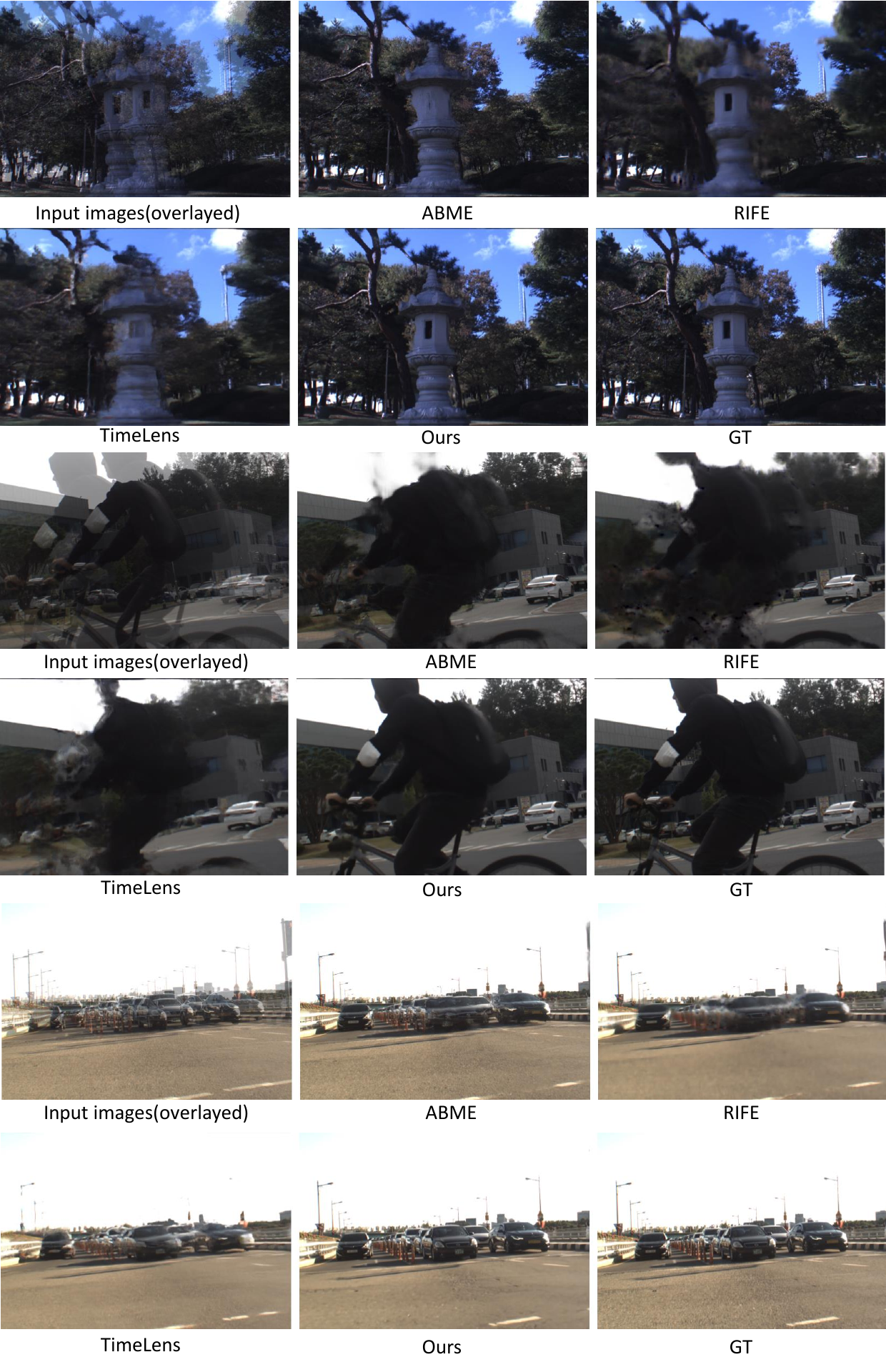} 
\vspace{-3mm}
\caption{Visual results on the ERF-X170FPS dataset. (Best viewed when zoomed in.)}
\label{supp:visual_res_erf6}
\end{figure*}

In Fig.~\ref{supp:visual_res_erf1}$\sim$Fig.~\ref{supp:visual_res_erf6}, we show more qualitative results of interpolated frames on the ERF-X170FPS dataset.
In the figure, we compare with state-of-the-art frame-based video frame interpolation methods, ABME~\cite{ABME}, RIFE~\cite{RIFE}, event-based video frame interpolation method, TimeLens~\cite{timelens}.
We confirm that our method significantly outperforms other frame- and event-based video frame interpolation methods.

\subsubsection{More Visual Results on GoPro~\cite{nah2017deep} dataset}
\begin{figure*}[!t]
\centering
\vspace{-3mm}
\includegraphics[width=0.91\linewidth]{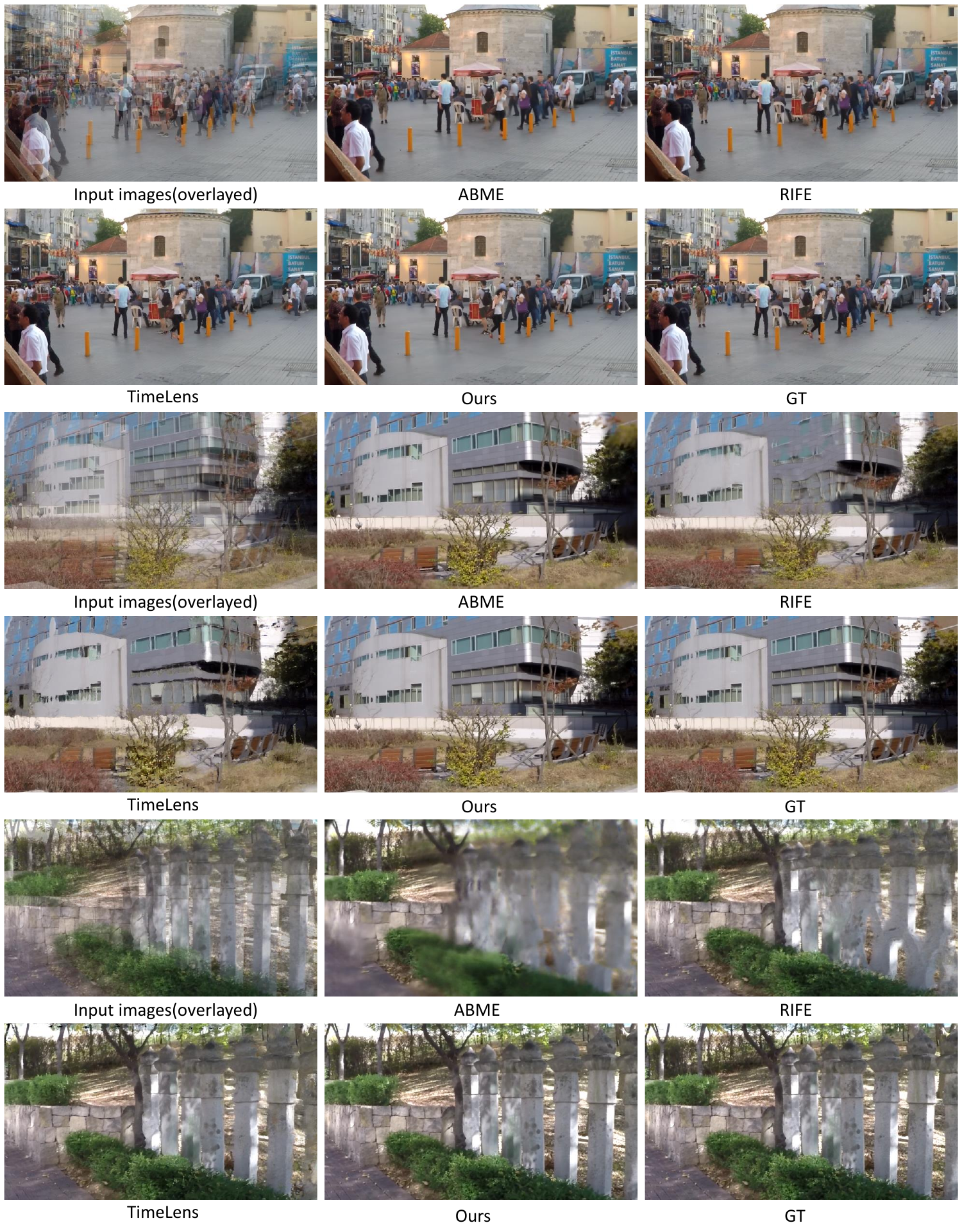} 
\vspace{-1mm}
\caption{Visual results on the GoPro dataset. (Best viewed when zoomed in.)}
\label{supp:visual_res_gopro}
\end{figure*}

In the Fig.~\ref{supp:visual_res_gopro}, we show more qualitative results on the GoPro dataset.
\subsubsection{More Visual Results on Adobe240fps~\cite{su2017deep} dataset}
\begin{figure*}[!t]
\centering
\vspace{-3mm}
\includegraphics[width=0.91\linewidth]{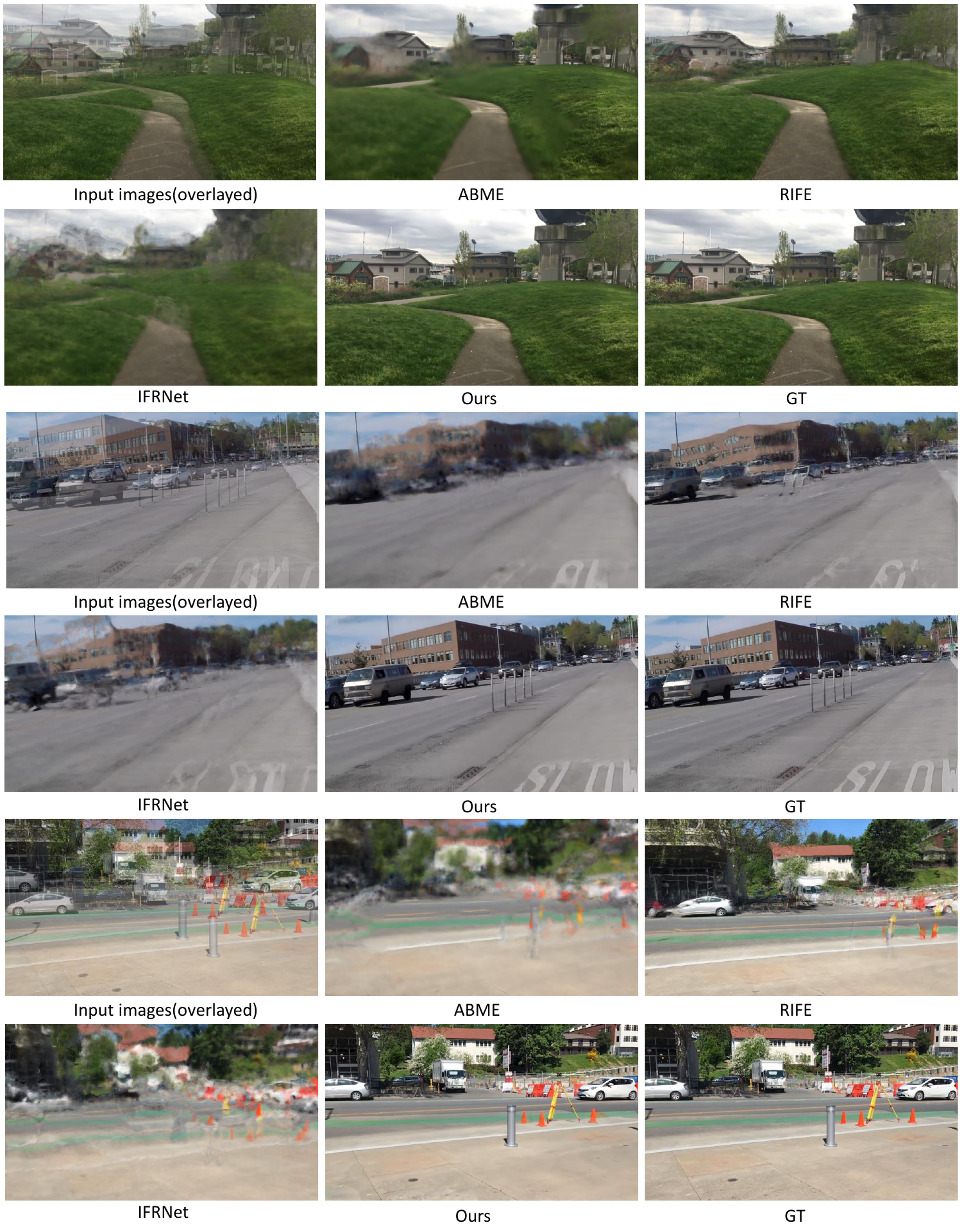} 
\vspace{-1mm}
\caption{Visual results on the Adobe240fps dataset. (Best viewed when zoomed in.)}
\label{supp:visual_res_adobe}
\end{figure*}

In the Fig.~\ref{supp:visual_res_adobe}, we show more qualitative results on the Adobe240fps dataset.

\begin{figure*}[!t]
\centering
\vspace{-3mm}
\includegraphics[width=0.84\linewidth]{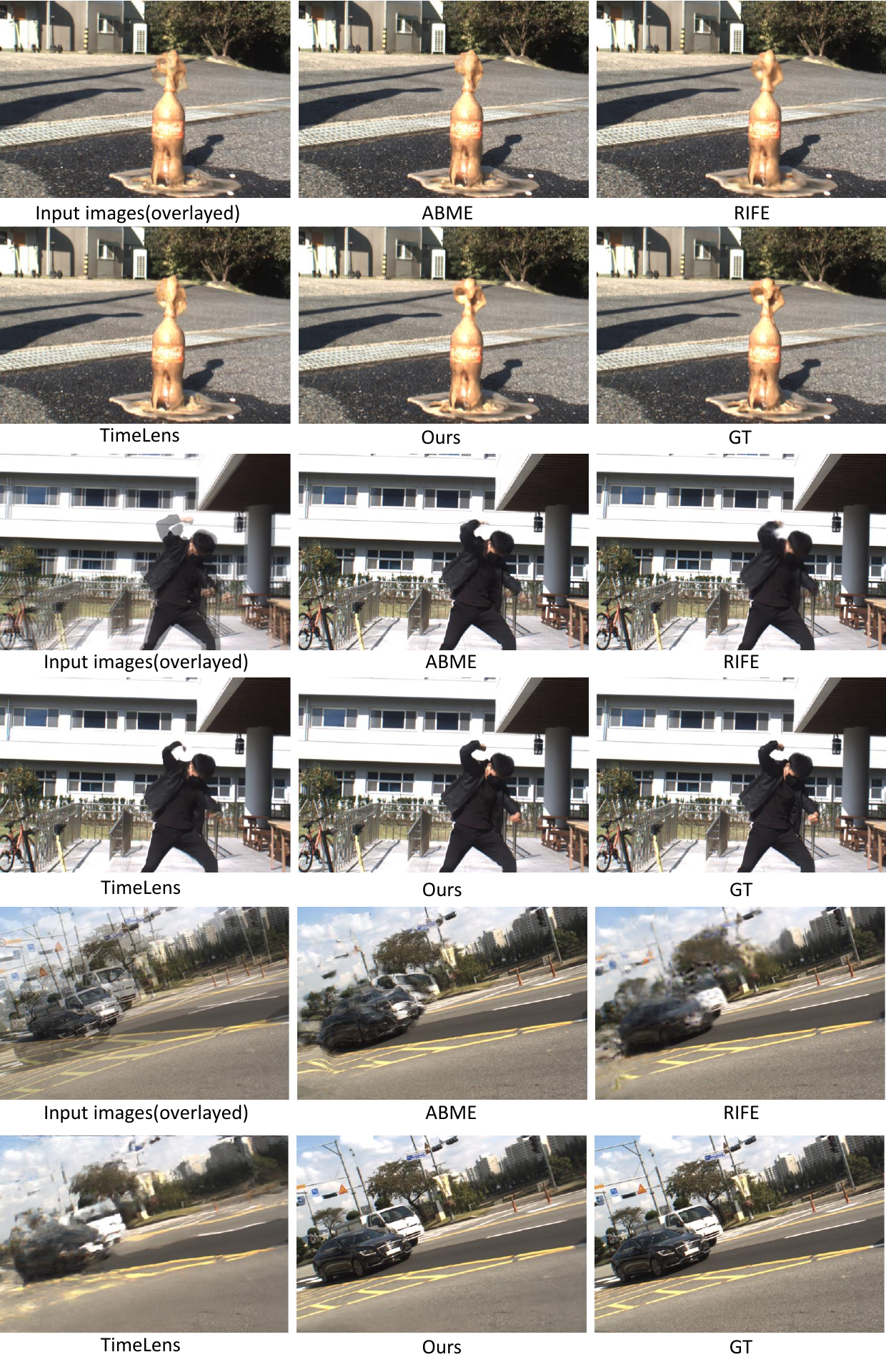} 
\vspace{-3mm}
\caption{Visual results on the ERF-X170FPS dataset. (Best viewed when zoomed in.)}
\label{supp:visual_res_erf2}
\end{figure*}

\begin{figure*}[!t]
\centering
\vspace{-3mm}
\includegraphics[width=0.84\linewidth]{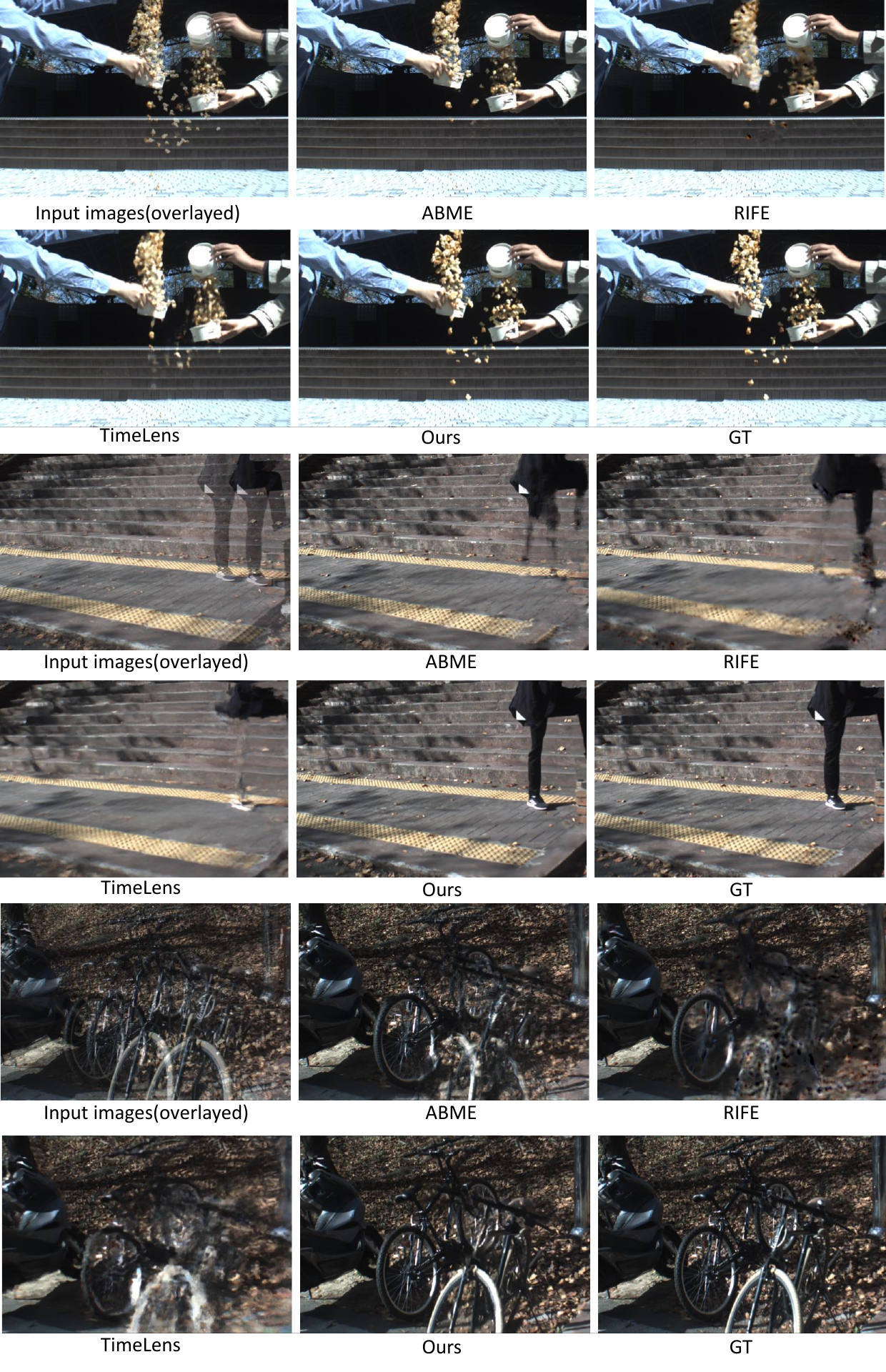} 
\vspace{-3mm}
\caption{Visual results on the ERF-X170FPS dataset. (Best viewed when zoomed in.)}
\label{supp:visual_res_erf3}
\end{figure*}

\begin{figure*}[!t]
\centering
\vspace{-3mm}
\includegraphics[width=0.84\linewidth]{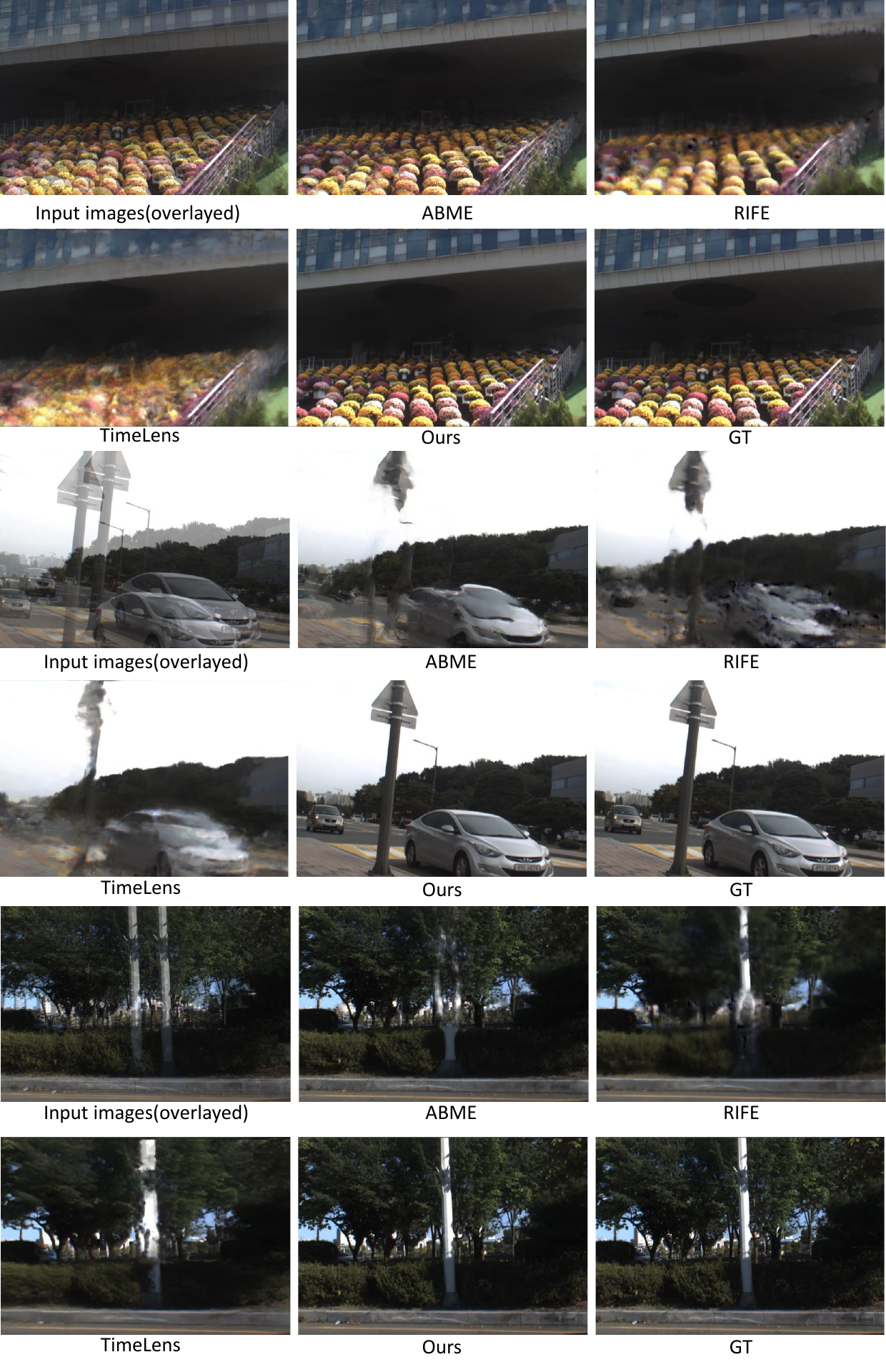} 
\vspace{-3mm}
\caption{Visual results on the ERF-X170FPS dataset. (Best viewed when zoomed in.)}
\label{supp:visual_res_erf4}
\end{figure*}

\begin{figure*}[!t]
\centering
\vspace{-3mm}
\includegraphics[width=0.84\linewidth]{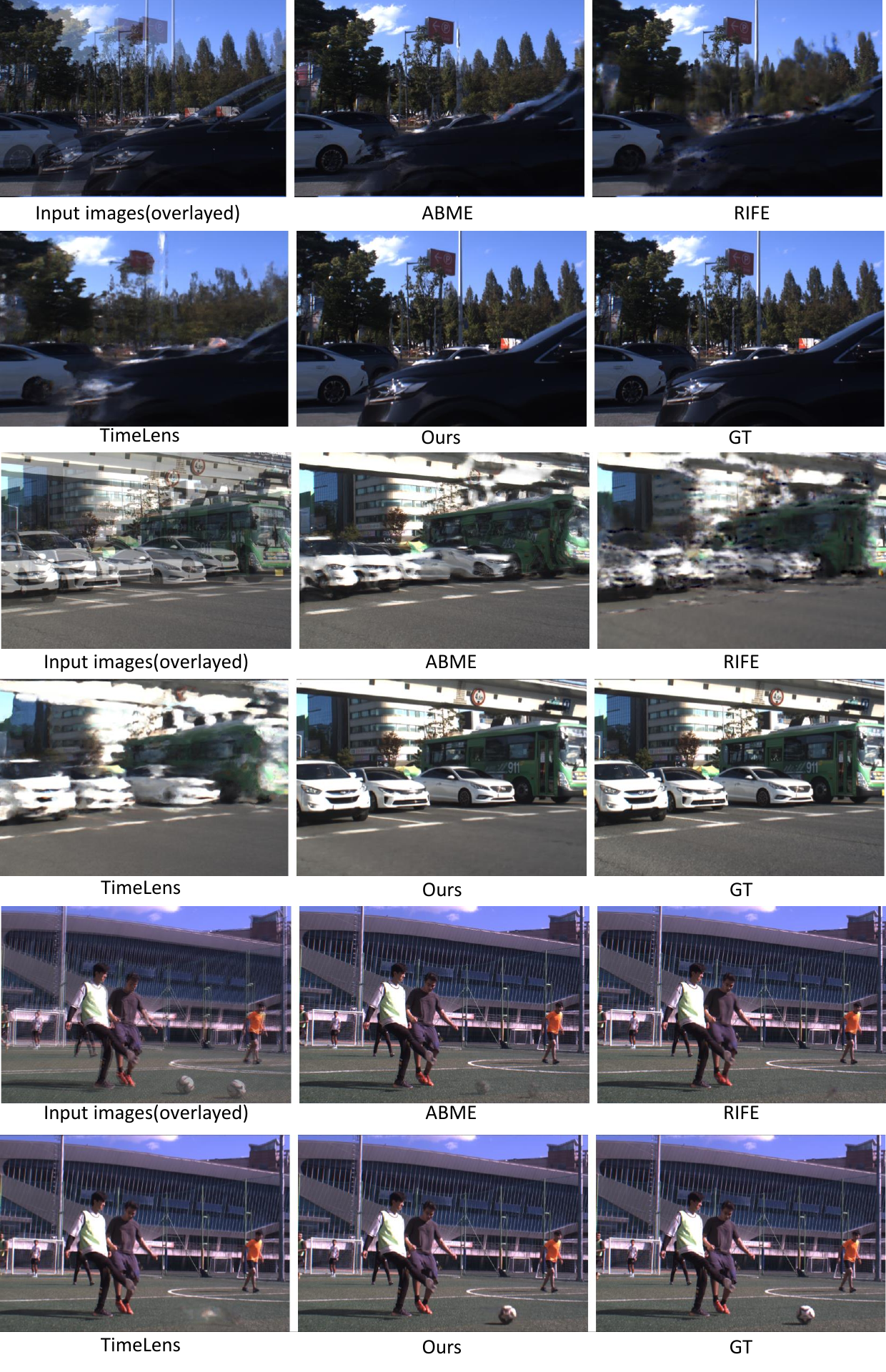} 
\vspace{-3mm}
\caption{Visual results on the ERF-X170FPS dataset. (Best viewed when zoomed in.)}
\label{supp:visual_res_erf5}
\end{figure*}

\end{document}